\documentclass[10pt,journal,compsoc]{IEEEtran}

\ifCLASSOPTIONcompsoc
  \usepackage[nocompress]{cite}
\else
  \usepackage{cite}
\fi

\usepackage{amsmath}
\usepackage{amsfonts}
\usepackage{graphicx}
\usepackage{multirow}
\ifCLASSINFOpdf
\else
\fi
\ifCLASSOPTIONcompsoc
\usepackage[caption=false,font=normalsize,labelfon
t=sf,textfont=sf]{subfig}
\else
\usepackage[caption=false,font=footnotesize]{subfi
g}
\fi

\usepackage[colorlinks,
            linkcolor=black,
            anchorcolor=black,
           citecolor=black,
            urlcolor=black
           ]{hyperref}

\begin{document}
%
\title{Shakeout: A New Approach to Regularized \\  Deep Neural Network Training}
%
%
%
%

\author{Guoliang~Kang,
        Jun~Li,
        and~Dacheng~Tao,~\IEEEmembership{Fellow,~IEEE}
\IEEEcompsocitemizethanks{\IEEEcompsocthanksitem Guoliang Kang, Jun Li are with Center of AI, Faculty of Engineering and Information Technology, University of Technology Sydney, Ultimo, NSW, Australia. D. Tao is with the UBTech Sydney Artificial Intelligence Institute and the School of Information Technologies in the Faculty of Engineering and Information Technologies at The University of Sydney, Darlington, NSW 2008, Australia.
\protect\\
E-mail: Guoliang.Kang@student.uts.edu.au, junjy007@googlemail.com, dacheng.tao@sydney.edu.au 
\protect\\
~\copyright 2018 IEEE. Personal use of this material is permitted. Permission from IEEE must be obtained for all other uses, in any current or future media, including reprinting/republishing this material for advertising or promotional purposes, creating new collective works, for resale or redistribution to servers or lists, or reuse of any copyrighted component of this work in other works.}
}

\IEEEtitleabstractindextext{%
\begin{abstract}
Recent years have witnessed the success of deep neural networks in
dealing with a plenty of practical problems. Dropout has played an essential role in many successful deep neural networks, by inducing regularization in the model training. In this paper,
we present a new regularized training approach: Shakeout. Instead of randomly
discarding units as Dropout does at the training stage, Shakeout randomly
chooses to enhance or reverse each unit's contribution to the
next layer. 
This minor modification of Dropout has the statistical trait: 
the regularizer induced by Shakeout adaptively combines $L_{0}$, $L_{1}$ and $L_{2}$ regularization terms. 
Our classification experiments with representative deep architectures on image datasets MNIST, CIFAR-10 and
ImageNet show that Shakeout deals with over-fitting effectively and
outperforms Dropout.  
We empirically
demonstrate that Shakeout leads to sparser weights under both unsupervised
and supervised settings. Shakeout also leads to the
grouping effect of the input units in a layer. Considering the weights in reflecting
the importance of connections, Shakeout is superior to Dropout, which
is valuable for the deep model compression. Moreover, we demonstrate that Shakeout can effectively reduce the instability of the training process of the deep architecture.
\end{abstract}

}

\maketitle

\IEEEdisplaynontitleabstractindextext

%
\IEEEpeerreviewmaketitle

\global\long\def\pr{P}
\global\long\def\loss{l}
\global\long\def\lsko{\loss_{\textrm{sko}}}
\global\long\def\lappox{l_{A}}
\global\long\def\w{\boldsymbol{w}}
\global\long\def\x{\boldsymbol{x}}
\global\long\def\r{\boldsymbol{r}}
\global\long\def\Regw{\mathrm{\pi(\boldsymbol{w})}}
\global\long\def\vy{\textrm{Var}[y|\x,\w]}
\global\long\def\pen{\left\Vert \x\circ(\w+c\boldsymbol{s})\right\Vert _{2}^{2}}
\global\long\def\thetaq{\theta_{j-}}
\global\long\def\thetaQ{\theta_{j+}}

\ifCLASSOPTIONcompsoc
\IEEEraisesectionheading{\section{Introduction}\label{sec:introduction}}
\else
\section{Introduction}
\label{sec:introduction}
\fi

%
%
%
%
\IEEEPARstart{D}{eep} neural networks have recently achieved impressive success in
a number of machine learning and pattern recognition tasks and been
under intensive research \cite{he2016identity, DBLP:conf/aaai/SzegedyIVA17, 7112511, 7346495, 7258387,wei2016hcp,jin2015deep}. Hierarchical neural networks have been known
for decades, and there are a number of essential factors contributing
to its recent rising, such as the availability of big data and powerful
computational resources. However, arguably the most important contributor
to the success of deep neural network is the discovery of efficient
training approaches \cite{hinton2006fast,bengio2009learning,6472238,vincent2008extracting,vincent2010stacked}.

A particular interesting advance in the training techniques is the
invention of Dropout \cite{hinton2012improving}. At the operational
level, Dropout adjusts the network evaluation step (feed-forward)
at the training stage, where a portion of units are randomly discarded.
The effect of this simple trick is impressive. Dropout enhances
the generalization performance of neural networks considerably, and
is behind many record-holders of widely recognized benchmarks \cite{krizhevsky2012imagenet,DBLP:conf/aaai/SzegedyIVA17,DBLP:conf/bmvc/ZagoruykoK16}.
The success has attracted much research attention, and found applications in a wider range of problems \cite{wager2013dropout,chen2014dropout,van2013learning}.
Theoretical research from the viewpoint of statistical learning has
pointed out the connections between Dropout and model
regularization, which is the de facto recipe of reducing
over-fitting for complex models in practical machine learning. For
example, Wager et al. \cite{wager2013dropout} showed
that for a generalized linear model (GLM), Dropout implicitly imposes
an adaptive $L_{2}$ regularizer of the network weights through
an estimation of the inverse diagonal Fisher information matrix.  

Sparsity is of vital importance in deep learning. It is straightforward
that through removing unimportant weights, deep neural networks perform
prediction faster. Additionally, it is expected to obtain better generalization
performance and reduce the number of examples needed in the training
stage \cite{lecun1989optimal}. Recently much evidence
has shown that the accuracy of a trained deep neural network will
not be severely affected by removing a majority of connections and 
many researchers focus on the deep model compression task \cite{DBLP:conf/icml/ChenWTWC15,
han2015learning,han2015deep,
denil2013predicting,ba2014deep,hinton2015distilling}.
One effective way of compression is to train a neural network, prune the connections and fine-tune the weights iteratively \cite{han2015learning,han2015deep}. However,
if we can cut the connections naturally via imposing sparsity-inducing penalties
in the training process of a deep neural network, the work-flow will
be greatly simplified.

In this paper, we propose a new regularized deep neural network training
approach: Shakeout, which is easy to implement: randomly choosing to
enhance or reverse each unit's contribution to the next layer in the training stage.
Note that Dropout can be considered as a special \textquotedblleft flat" case of our approach: randomly keeping (enhance factor is $1$) or discarding (reverse factor is $0$) each unit's contribution to the next layer.
Shakeout enriches the regularization effect.
In theory, we prove that it adaptively combines $L_{0}$, $L_{1}$ and $L_{2}$
regularization terms. $L_{0}$ and $L_{1}$ regularization terms are known as sparsity-inducing penalties.
The combination of sparsity-inducing
penalty and $L_{2}$ penalty of the model parameters has shown to
be effective in statistical learning: the Elastic Net \cite{zou2005regularization}
has the desirable properties of producing sparse models while maintaining
the grouping effect of the weights of the model.
Because of the randomly
\textquotedblleft shaking" process and the regularization
characteristic pushing network weights to zero, our new approach is named \textquotedblleft Shakeout". 

As discussed above, it is expected to obtain much sparser weights
using Shakeout than using Dropout because of the combination of $L_{0}$
and $L_{1}$ regularization terms induced in the training stage. We apply
Shakeout on one-hidden-layer autoencoder and obtain much sparser
weights than that resulted by Dropout. To show the regularization
effect on the classification tasks, we conduct the experiments
on image datasets including MNIST, CIFAR-10 and ImageNet with the
representative deep neural network architectures. In our experiments
we find that by using Shakeout, the trained deep neural networks always 
outperform those by using Dropout, especially when the data is scarce.
Besides
the fact that Shakeout leads to much sparser weights, we also empirically find that 
it groups the input units of a layer. Due to the induced $L_{0}$ and $L_{1}$ regularization terms,
Shakeout can result in the weights reflecting the importance
of the connections between units, which is meaningful for conducting
compression. Moreover, we demonstrate that Shakeout can effectively reduce the instability of the training process of the deep architecture.

This journal paper extends our previous work \cite{kang2016shakeout} theoretically and experimentally. The main extensions are listed as follows:
1) we derive the analytical formula for the regularizer induced by Shakeout in the context of GLM and prove several important properties; 
2) we conduct experiments using Wide Residual Network \cite{DBLP:conf/bmvc/ZagoruykoK16} on CIFAR-10 to show Shakeout outperforms Dropout and standard back-propagation in promoting the generalization performance of a much deeper architecture;
3) we conduct experiments using AlexNet \cite{krizhevsky2012imagenet} on ImageNet dataset with Shakeout and Dropout. Shakeout obtains comparable classification performance to Dropout, but with superior regularization effect;
4) we illustrate that Shakeout can effectively reduce the instability of the training process of the deep architecture.
Moreover, we provide a much clearer and detailed description of Shakeout, derive the forward-backward update rule for deep convolutional neural networks with Shakeout, and give several recommendations to help the practitioners make full use of Shakeout.

In the rest of the paper, we give a review
about the related work in Section 2. 
Section 3 presents Shakeout in detail, along with theoretical analysis of the regularization effect induced by Shakeout.
In Section 4, we first demonstrate the regularization effect of Shakeout 
on the autoencoder model. The classification experiments
on MNIST , CIFAR-10 and ImageNet illustrate that Shakeout outperforms
Dropout considering the generalization performance, the regularization effect on the weights, and the stabilization effect on the training process of the deep architecture. Finally, we give some recommendations for the practitioners
to make full use of Shakeout.

\section{Related Work}
Deep neural networks have shown their success in a wide variety of applications.
One of the key factors contributes to this success is the creation of powerful training techniques. 
The representative power of the network becomes stronger as the architecture gets deeper \cite{bengio2009learning}. However, millions of parameters make deep neural networks easily over-fit.  
Regularization \cite{erhan2010does,wager2013dropout} is an effective way 
to obtain a model that generalizes well.
There exist many approaches to regularize the training of deep neural networks, like weight decay \cite{moody1995simple}, early stopping \cite{prechelt1998automatic}, etc. Shakeout belongs to the family of regularized training techniques. 

Among these regularization techniques, our work is closely related to Dropout \cite{hinton2012improving}.
Many subsequent works were devised to improve the performance of Dropout \cite{wan2013regularization,ba2013adaptive,li2016improved}.
The underlying reason why Dropout improves performance has also attracted
the interest of many researchers. Evidence has shown that Dropout
may work because of its good approximation to model averaging and
regularization on the network weights
\cite{srivastava2014dropout, warde2013empirical,baldi2013understanding}. Srivastava \cite{srivastava2014dropout}
and Warde-Farley \cite{warde2013empirical} exhibited through experiments
that the weight scaling approximation is an accurate alternative for the geometric
mean over all possible sub-networks. Gal et al. \cite{DBLP:conf/icml/GalG16} claimed that 
training the deep neural network with Dropout is equivalent to
performing variational inference in a deep Gaussian Process.
Dropout can also be regarded as a
way of adding noise into the neural network. By marginalizing the
noise, Srivastava \cite{srivastava2014dropout} proved for linear
regression that the deterministic version of Dropout is equivalent
to adding an adaptive $L_{2}$ regularization on the weights. Furthermore,
Wager \cite{wager2013dropout} extended the conclusion to generalized
linear models (GLMs) using a quadratic approximation to the induced
regularizer. 
The inductive bias of Dropout was studied by Helmbold et al. \cite{helmbold2015inductive} to illustrate the properties of the regularizer induced
by Dropout further. 
In terms of implicitly inducing regularizer of the network weights, Shakeout can be viewed as a generalization of Dropout. It enriches the regularization effect of Dropout, i.e. besides the $L_{2}$ regularization term, it also induces the $L_{0}$ and $L_{1}$ regularization terms, which may lead to sparse weights of the model.

Due to the implicitly induced $L_{0}$ and $L_{1}$ regularization terms, Shakeout is also related to  sparsity-inducing approaches. Olshausen et al. \cite{olshausen1997sparse}
introduced the concept of sparsity in computational neuroscience and proposed
the sparse coding method in the visual system. In machine learning,
the sparsity constraint enables a model to capture the implicit statistical
data structure, performs feature selection and regularization, compresses
the data at a low loss of the accuracy, and helps us to better understand
our models and explain the obtained results. Sparsity is one of the key factors underlying many successful deep neural network architectures \cite{lecun1998gradient,szegedy2015going,szegedy2016rethinking,DBLP:conf/aaai/SzegedyIVA17}
and training algorithms \cite{boureau2008sparse}\cite{goodfellow2012spike}.
A Convolutional neural network
is much sparser than the fully-connected one, which results from
the concept of local receptive field \cite{lecun1998gradient}. Sparsity
has been a design principle and motivation for Inception-series
models \cite{szegedy2015going,szegedy2016rethinking,DBLP:conf/aaai/SzegedyIVA17}.
Besides working as the heuristic principle of designing a deep
architecture, sparsity often works as a penalty induced to regularize the training process of a deep neural network. There exist two kinds of sparsity penalties in deep neural
networks, which lead to the activity sparsity \cite{boureau2008sparse}\cite{goodfellow2012spike} 
and the connectivity sparsity
\cite{thom2013sparse} respectively.
The difference between Shakeout and these sparsity-inducing approaches is that for Shakeout, the sparsity is induced through simple stochastic operations rather than manually designed architectures or explicit norm-based penalties. This implicit way enables Shakeout to be easily optimized by stochastic gradient descent (SGD) ${-}$ the representative approach for the optimization of a deep neural network. 

\section{Shakeout}
Shakeout applies on the weights in a linear module.
The linear module, i.e. weighted sum, 
\begin{align}
\theta & =\sum_{j=1}^{p}w_{j}x_{j}\label{eq:w-sum}
\end{align}
is arguably the most widely adopted component in data
models. For example, the variables $x_{1}$, $x_{2}$, $\dots$,
$x_{p}$ can be input attributes of a model, e.g. the extracted features for a GLM, or the intermediate outputs of earlier processing steps, e.g. the activations of the hidden units in a multilayer artificial neural network. Shakeout \textit{randomly}
modifies the computation in Eq. (\ref{eq:w-sum}). Specifically, Shakeout
can be realized by randomly modifying the weights

\textbf{\textit{Step 1}}: Draw $r_{j}$, where $\begin{cases}
\pr(r_{j}=0) & =\tau\\
\pr(r_{j}=\frac{1}{1-\tau}) & =1-\tau
\end{cases}$ .

\textbf{\textit{Step 2}}: Adjust the weight according to $r_{j}$,
\begin{multline*}
\begin{cases}
\tilde{w}_{j}\leftarrow-c s_j, & \ \textrm{if }r_{j}=0\qquad\,\,\,\textrm{\textrm{(A)}}\\
\tilde{w}_{j}\leftarrow(w_{j}+c\tau s_j)/(1-\tau) & \ \textrm{otherwise}\qquad\textrm{(B)}
\end{cases}
\end{multline*}
where $s_{j}=\textrm{sgn}(w_{j})$ takes $\pm1$ depending on the sign of
$w_{j}$ or takes 0 if $w_{j}=0$. As shown above, Shakeout chooses
(randomly by drawing $r$) between two fundamentally different ways
to modify the weights. Modification (A) is to set the weights to
constant magnitudes, despite their original values except for the signs
(to be opposite to the original ones). Modification (B) updates the
weights by a factor $(1-\tau)^{-1}$ and a bias depending on the signs.
Note both (A) and (B) preserve zero values of the weights, i.e. if $w_{j}=0$
then $\tilde{w}_{j}=0$ with probability 1. Let $\tilde{\theta}=\tilde{\w}^{T}\x$,
and Shakeout leaves $\theta$ unbiased, i.e.
$\mathbb{E}[\tilde{\theta}]=\theta$. 
The hyper-parameters $\tau\in(0,1)$ and $c\in(0,+\infty)$ configure the property of Shakeout.

\begin{figure}[!t]
\centering
\includegraphics[bb=240bp 70bp 550bp 600bp,scale=0.3]{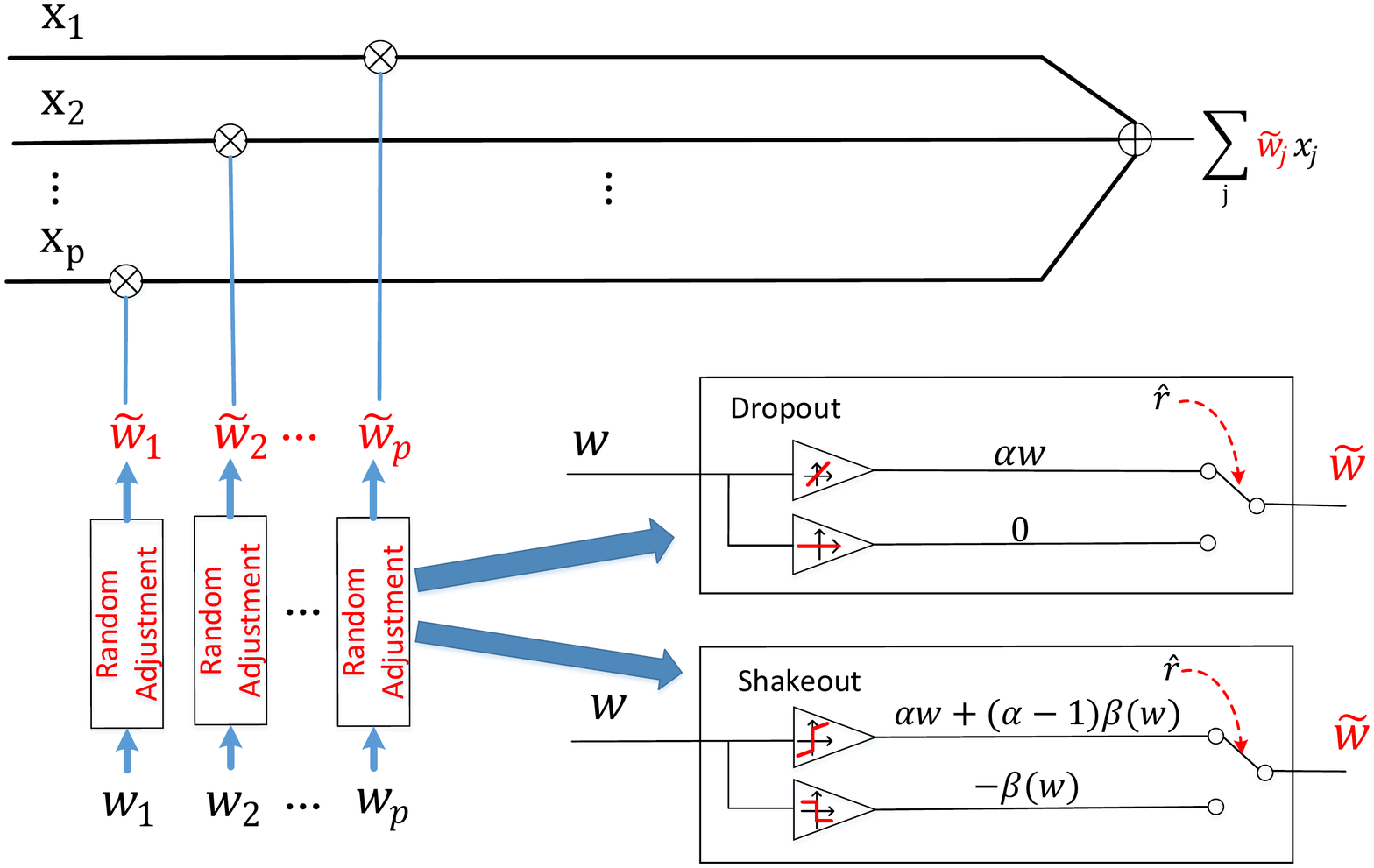}
\protect\caption{Comparison between Shakeout and
Dropout operations. This figure shows how Shakeout and Dropout are
applied to the weights in a linear module. In the original linear
module, the output is the summation of the inputs $\x$ weighted
by $\w$, while for Dropout and Shakeout, the weights $\w$
are first randomly modified. In detail, a random switch $\hat{r}$
controls how each $w$ is modified. The manipulation of $w$ is illustrated
within the amplifier icons (the red curves, best seen with colors).
The coefficients are $\alpha=1/(1-\tau)$ and $\beta(w)=cs(w)$, where
$s(w)$ extracts the sign of $w$ and $c>0$, $\tau\in[0,1]$. Note
the sign of $\beta(w)$ is always the same as that of $w$.
The magnitudes of coefficients $\alpha$ and $\beta(w)$ are determined
by the Shakeout hyper-parameters $\tau$ and $c$. Dropout can be
viewed as a special case of Shakeout when $c=0$ because $\beta(w)$
is zero at this circumstance.}
\label{fig:The-Shakeout-operations}
\end{figure}

Shakeout is naturally connected to the widely adopted operation of
Dropout \cite{hinton2012improving,srivastava2014dropout}. We will
show that Shakeout has regularization effect on model training
similar to but beyond what is induced by Dropout. From an operational
point of view, Fig. \ref{fig:The-Shakeout-operations} compares Shakeout
and Dropout. Note that Shakeout includes Dropout as a special case
when the hyper-parameter $c$ in Shakeout is set to zero.

When applied at the training stage, Shakeout alters the objective
$-$ the quantity to be minimized $-$ by adjusting the weights. In
particular, we will show that Shakeout (with expectation over the
random switch) induces a regularization term effectively penalizing
the magnitudes of the weights and leading to sparse weights. Shakeout
is an approach designed for helping model training, when the models are
trained and deployed, one should relieve the disturbance to allow
the model work with its full capacity, i.e. 
we adopt the resulting network without any modification of the weights at the test stage.

\subsection{\label{sub:Shakeout-for-GLM}Regularization Effect of Shakeout}
Shakeout randomly modifies the weights in a linear module, and thus
can be regarded as injecting noise into each variable $x_{j}$, i.e.
$x_{j}$ is randomly scaled by $\gamma_{j}$: $\tilde{x}_{j}=\gamma_{j}x_{j}$. 
Note that $\gamma_{j}=r_{j}+\frac{c(r_{j}-1)}{|w_j|}$, the modification of $x_{j}$ is actually determined by the random switch ${r_{j}}$.
Shakeout randomly
chooses to enhance (i.e. when  $r_{j}=\frac{1}{1-\tau}$, $\gamma_{j}>\frac{1}{1-\tau}$)
or reverse (i.e. when  $r_{j}=0$, $\gamma_{j}<0$) each original variable $x_{j}$'s contribution to the output at the training stage 
(see Fig. \ref{fig:The-Shakeout-operations}). However, the
expectation of $\tilde{x}_{j}$ over the noise remains unbiased, i.e. $\mathbb{E}_{r_{j}}[\tilde{x}_{j}]=x_{j}$.

It is well-known that injecting artificial noise into the
input features will regularize the training objective \cite{wager2013dropout,rifai2011adding,bishop1995training},
i.e. $\mathbb{E}_{\r}[\ell(\boldsymbol{w},\tilde{\boldsymbol{x}},y)]=\ell(\boldsymbol{w},\boldsymbol{x},y)+\pi(\boldsymbol{w})$,
where $\tilde{\boldsymbol{x}}$ is the feature vector randomly modified by
the noise induced by $\r$. 
The regularization
term $\pi(\boldsymbol{w})$ is determined by the characteristic of
the noise. For example, Wager et al.\cite{wager2013dropout} showed
that Dropout, corresponding to inducing blackout noise to the features,
helps introduce an adaptive $L_{2}$ penalty on $\boldsymbol{w}$.
In this section we illustrate how Shakeout helps regularize model
parameters $\boldsymbol{w}$ using an example of GLMs. 

Formally, a GLM is a probabilistic model of predicting target $y$ given
features $\x=[x_{1},\dots,x_{p}]$, in terms of the weighted sum in Eq. (\ref{eq:w-sum}):
\begin{eqnarray}
\pr(y|\boldsymbol{x},\boldsymbol{w}) & = & h(y)g(\theta)e^{\theta y}\label{eq:GLM-eq}\\
\theta & = & \w^{T}\x\nonumber 
\end{eqnarray}
With different $h(\cdot)$ and $g(\cdot)$ functions, GLM can be specialized
to various useful models or modules, such as logistic regression model
or a layer in a feed-forward neural network. However, roughly
speaking, the essence of a GLM is similar to that of a standard linear
model which aims to find weights $w_{1},\dots,w_{p}$ so that $\theta=\w^{T}\x$
aligns with $y$ (functions $h(\cdot)$ and $g(\cdot)$ are independent
of $\w$ and $y$ respectively). The loss function of a GLM
with respect to $\w$ is defined as
\begin{align}
\loss(\w,\x,y) & =-\theta y+A(\theta)\label{eq:glm-loss}\\
A(\theta) & =-\ln[g(\theta)]
\end{align}
The loss (\ref{eq:glm-loss}) is the negative logarithm of probability
(\ref{eq:GLM-eq}), where we keep only terms relevant to $\w$.

Let the loss with Shakeout be 
\begin{equation}
\lsko(\w,\x,y,\r):=\loss(\w,\tilde{\x},y)\label{eq:shakeout-loss}
\end{equation}
where $\r=[r_{1},\dots,r_{p}]^{T}$, and $\tilde{\x}=[\tilde{x}_{1},\dots,\tilde{x}_{p}]^{T}$
represents the features randomly modified with $\r$. 

Taking expectation over 
$\r$, the loss with Shakeout becomes
\[
\mathbb{E}_{\r}[\lsko(\w,\x,y,\r)]=\loss(\w,\x,y)+\Regw
\]
where
\begin{eqnarray}
\mathrm{\Regw} & = & \mathbb{E}_{\r}[A(\tilde{\theta})-A(\theta)]\nonumber\\
 & = & \sum_{k=1}^{\infty}\frac{1}{k!}A^{(k)}(\theta)\mathbb{E}[(\tilde{\theta}-\theta)^{k}]\label{eq:full-form-regw}
\end{eqnarray}
is named \textit{Shakeout regularizer}.
Note that if $A(\theta)$ is $k$-th order derivable, let the $k^{'}$
order derivative $A^{(k^{'})}(\theta)=0$ where $k^{'}>k$, to make
the denotation simple.

\newtheorem{theorem}{Theorem}
\begin{theorem}
\label{thm:shakeout-reg}Let $q_{j}=x_{j}(w_{j}+cs_{j})$, $\thetaq=\theta-q_{j}$
and $\thetaQ=\theta+\frac{\tau}{1-\tau}q_{j}$, then Shakeout
regularizer $\Regw$ is
\begin{equation}
\Regw=\tau\sum_{j=1}^{p}A(\thetaq)+(1-\tau)\sum_{j=1}^{p}A(\thetaQ)-pA(\theta)\label{eq:shakeout-reg-accurate}
\end{equation}
\end{theorem}
\begin{IEEEproof}
Note that $\tilde{\theta}-\theta=\sum_{j=1}^{p}q_{j}(r_{j}-1)$, then
for Eq. (\ref{eq:full-form-regw})
\begin{eqnarray*}
\mathbb{E}[(\tilde{\theta}-\theta)^{k}] & = & \sum_{j_{1}=1}^{p}\sum_{j_{2}=1}^{p}\cdots\sum_{j_{k}=1}^{p}\prod_{m=1}^{k}q_{j_{m}}\mathbb{E}[\prod_{m=1}^{k}(r_{j_{m}}-1)]
\end{eqnarray*}
Because arbitrary two random variables $r_{j_{m_{1}}}$, $r_{j_{m_{2}}}$ are independent unless $j_{m_{1}}=j_{m_{2}}$ and $\forall r_{j_{m}}$, 
$\mathbb{E}[r_{j_{m}}-1]=0$, then
\begin{eqnarray*}
\mathbb{E}[(\tilde{\theta}-\theta)^{k}] & = & \sum_{j=1}^{p}q_{j}^{k}\mathbb{E}[(r_{j}-1)^{k}]\\
 & = & \tau\sum_{j=1}^{p}(-q_{j})^{k}+(1-\tau)\sum_{j=1}^{p}(\frac{\tau}{1-\tau}q_{j})^{k}
\end{eqnarray*}
Then
\begin{eqnarray*}
\Regw & = & \tau\sum_{j=1}^{p}\sum_{k=1}^{\infty}\frac{1}{k!}A^{(k)}(\theta)(-q_{j})^{k}\\
 &  & +(1-\tau)\sum_{j=1}^{p}\sum_{k=1}^{\infty}\frac{1}{k!}A^{(k)}(\theta)(\frac{\tau}{1-\tau}q_{j})^{k}
\end{eqnarray*}
Further, let $\thetaq=\theta-q_{j}$, $\thetaQ=\theta+\frac{\tau}{1-\tau}q_{j}$,
$\Regw$ becomes
\[
\Regw=\tau\sum_{j=1}^{p}A(\thetaq)+(1-\tau)\sum_{j=1}^{p}A(\thetaQ)-pA(\theta)
\]
The theorem is proved.
\end{IEEEproof}
We illustrate several properties of Shakeout regularizer
based on Eq. (\ref{eq:shakeout-reg-accurate}). The proof of the following
propositions can be found in the appendices.
\newtheorem{prop}{Proposition}
\begin{prop}
$\pi(\boldsymbol{0})=0$
\end{prop}
\begin{prop}
\label{prop:neq0}
If $A(\theta)$ is convex, $\Regw\geq0$.
\end{prop}
\begin{prop}
\label{prop:tau-and-c}Suppose $\exists j$, $x_{j}w_{j}\neq0$. If $A(\theta)$ is convex, $\Regw$ monotonically
increases with $\tau$. If $A^{''}(\theta)>0$,
$\Regw$ monotonically increases with $c$.
\end{prop}
Proposition \ref{prop:tau-and-c} implies that the hyper-parameters
$\tau$ and $c$ relate to the strength of the regularization effect.
It is reasonable because higher $\tau$ or $c$ means the noise injected
into the features $\x$ has larger variance.

\begin{figure*}[!t]
\centering
\subfloat[Shakeout regularizer: $\tau=0.3$, $c=0.78$] {\includegraphics[bb=0bp 160bp 595bp 642bp,scale=0.3]{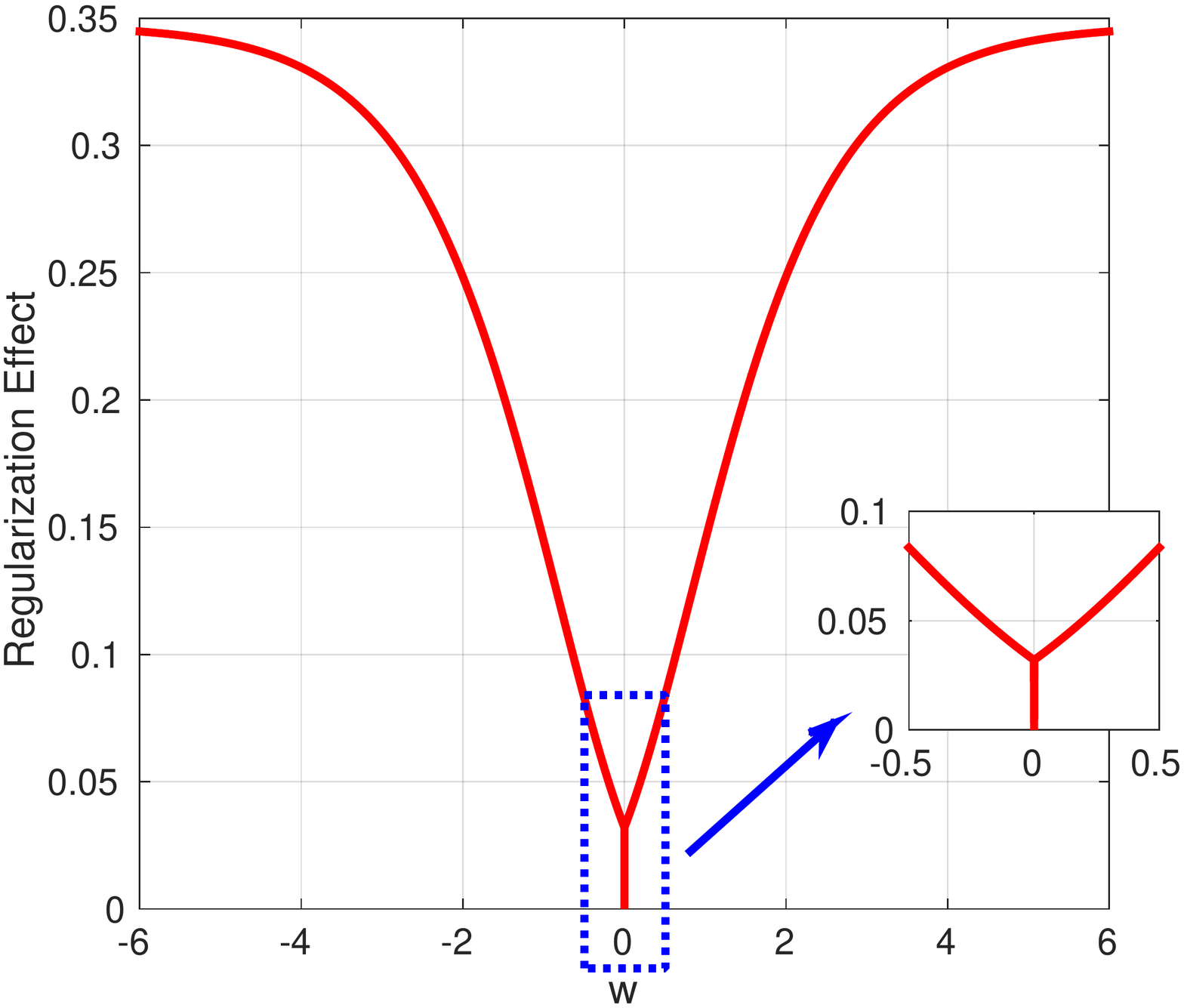}
\label{fig:comparision-with-reg-approx-a}}
\hfil
\subfloat[Dropout regularizer: $\tau=0.5$] {\includegraphics[bb=0bp 160bp 595bp 642bp,scale=0.3]{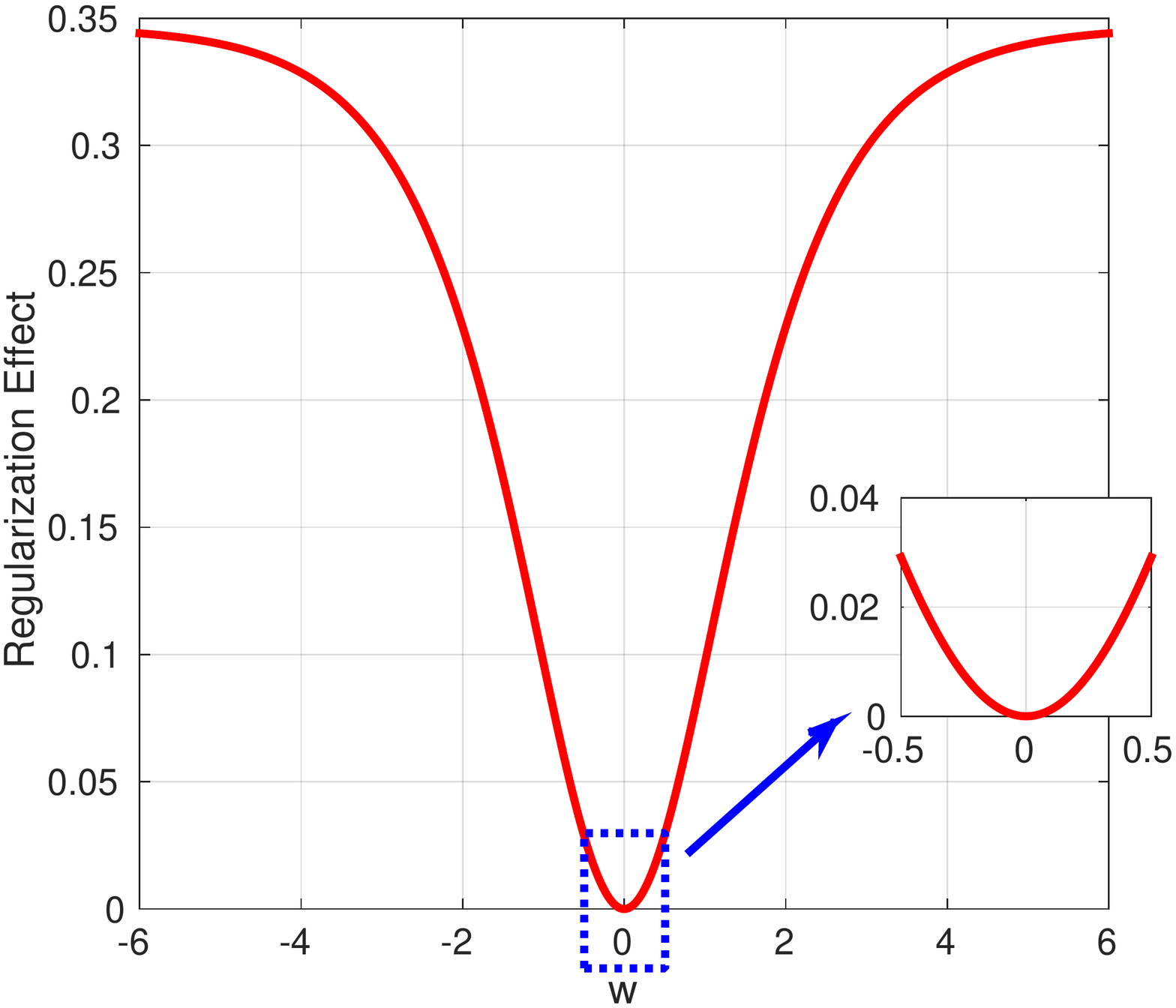}
\label{fig:comparision-with-reg-approx-b}}

\protect\caption{Regularization effect as a
function of a single weight when other weights are fixed to zeros
for logistic regression model. The corresponding feature $x$ is fixed
at 1.}
\label{fig:comparision-with-reg-approx}
\end{figure*}

\begin{prop}
\label{prop:sigle-w}Suppose \textit{i}) $\forall j\neq j^{'}$, $x_{j}w_{j}=0$,
and
\textit{ii}) $x_{j^{'}}\neq0$.

Then

\textit{i}) if $A^{''}(\theta)>0$,
$\begin{cases}
\frac{\partial\Regw}{\partial w_{j'}}>0, & \textrm{when}\ w_{j'}>0\\
\frac{\partial\Regw}{\partial w_{j'}}<0, & \textrm{when}\ w_{j'}<0
\end{cases}$

\textit{ii}) if $\lim_{|\theta|\rightarrow\infty}A^{''}(\theta)=0$,
$\lim_{|w_{j'}|\rightarrow\infty}\frac{\partial\Regw}{\partial w_{j'}}=0$
\end{prop}
Proposition \ref{prop:sigle-w} implies that under certain conditions, 
starting from a zero weight vector, Shakeout regularizer penalizes the magnitude of
$w_{j^{'}}$ and its regularization effect is bounded by a constant value. For example, for
logistic regression, $\Regw\leq\tau\ln(1+\exp(c|x_{j^{'}}|))$, which
is illustrated in Fig. \ref{fig:comparision-with-reg-approx}. This
bounded property has been proved to be useful: capped-norm \cite{DBLP:conf/ijcai/JiangNH15}
is more robust to outliers than the traditional $L_{1}$ or $L_{2}$
norm.

Based on the Eq. (\ref{eq:shakeout-reg-accurate}), the specific formulas
for the representative GLM models can be derived:

\textit{i}) Linear regression: $A(\theta)=\frac{1}{2}\theta^{2}$,
then
\[
\Regw=\frac{\tau}{2(1-\tau)}\pen
\]
where $\circ$ denotes the element-wise product and the $\pen$ term
can be decomposed into the summation of three components 
\begin{equation}
\sum_{j=1}^{p}x_{j}^{2}w_{j}^{2}+2c\sum_{j=1}^{p}x_{j}^{2}|w_{j}|+c^{2}\sum_{j=1}^{p}x_{j}^{2}\boldsymbol{1}_{w_{j}\neq0}[w_{j}]\label{eq:pen-decomp}
\end{equation}
where $\boldsymbol{1}_{w_{j}\neq0}[w_{j}]$ is an indicator function which satisfies
$\boldsymbol{1}_{w_{j}\neq0}[w_{j}]=\begin{cases}
1 & w_{j}\neq0\\
0 & w_{j}=0
\end{cases}$. This decomposition implies that Shakeout regularizer penalizes the
combination of $L_{0}$-norm, $L_{1}$-norm and $L_{2}$-norm of the weights
after scaling them with the square of corresponding features. The
$L_{0}$ and $L_{1}$ regularization terms can lead to sparse weights.

\begin{figure*}
\centering
\includegraphics[bb=100bp 150bp 595bp 642bp,scale=0.35]{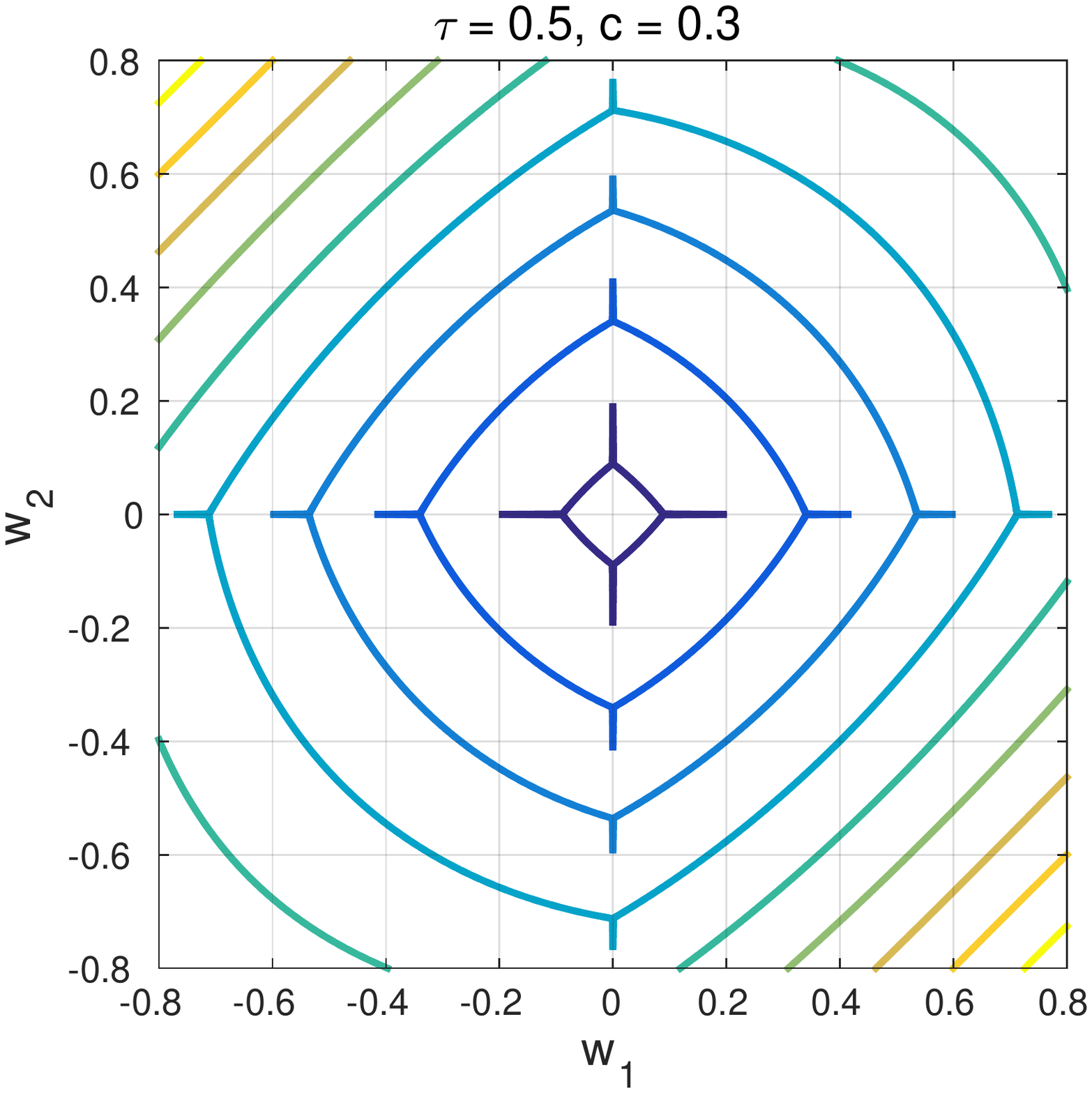}
\includegraphics[bb=70bp 150bp 495bp 642bp,scale=0.35]{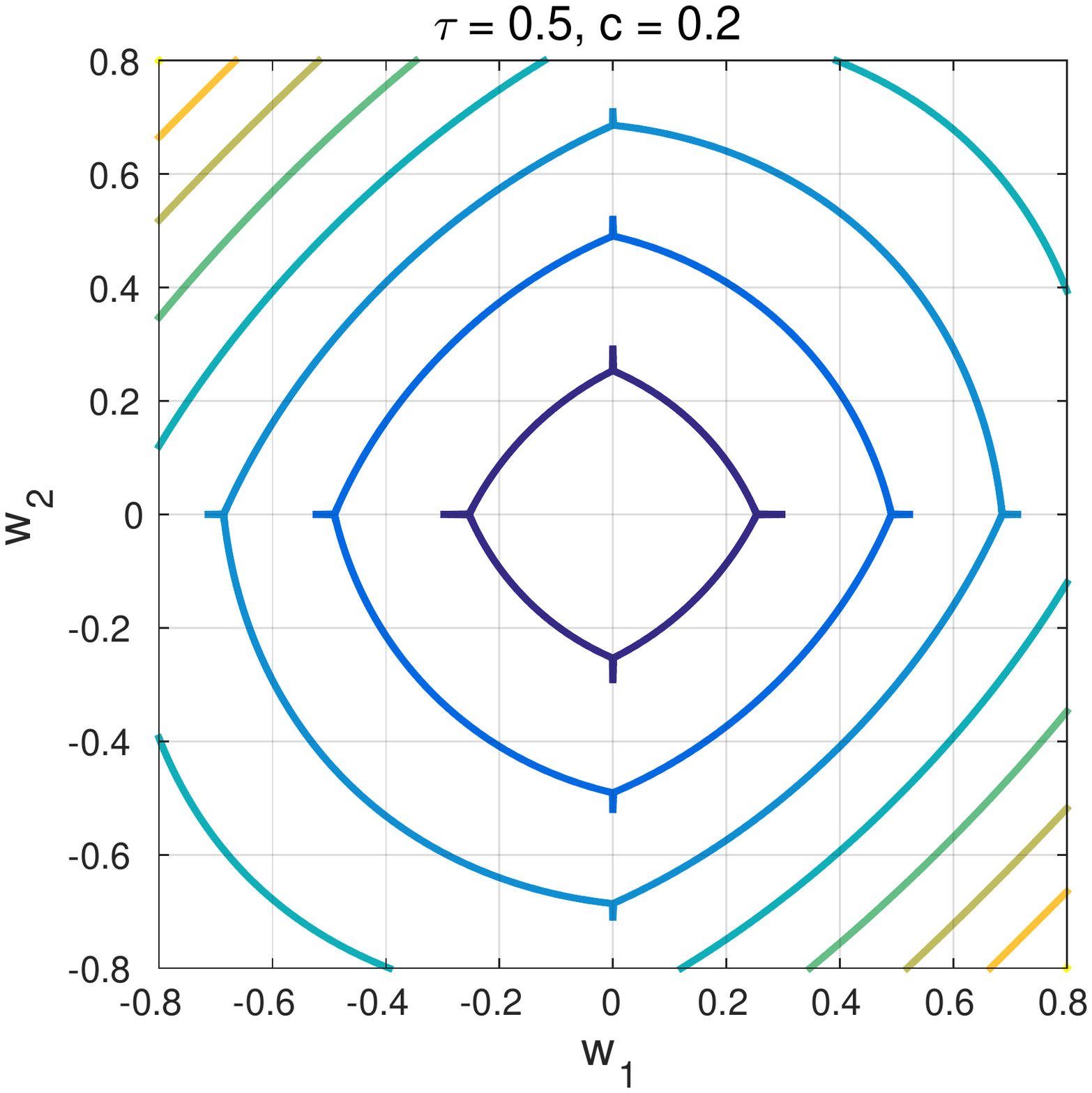}

\centering
\includegraphics[bb=100bp 180bp 595bp 642bp,scale=0.35]{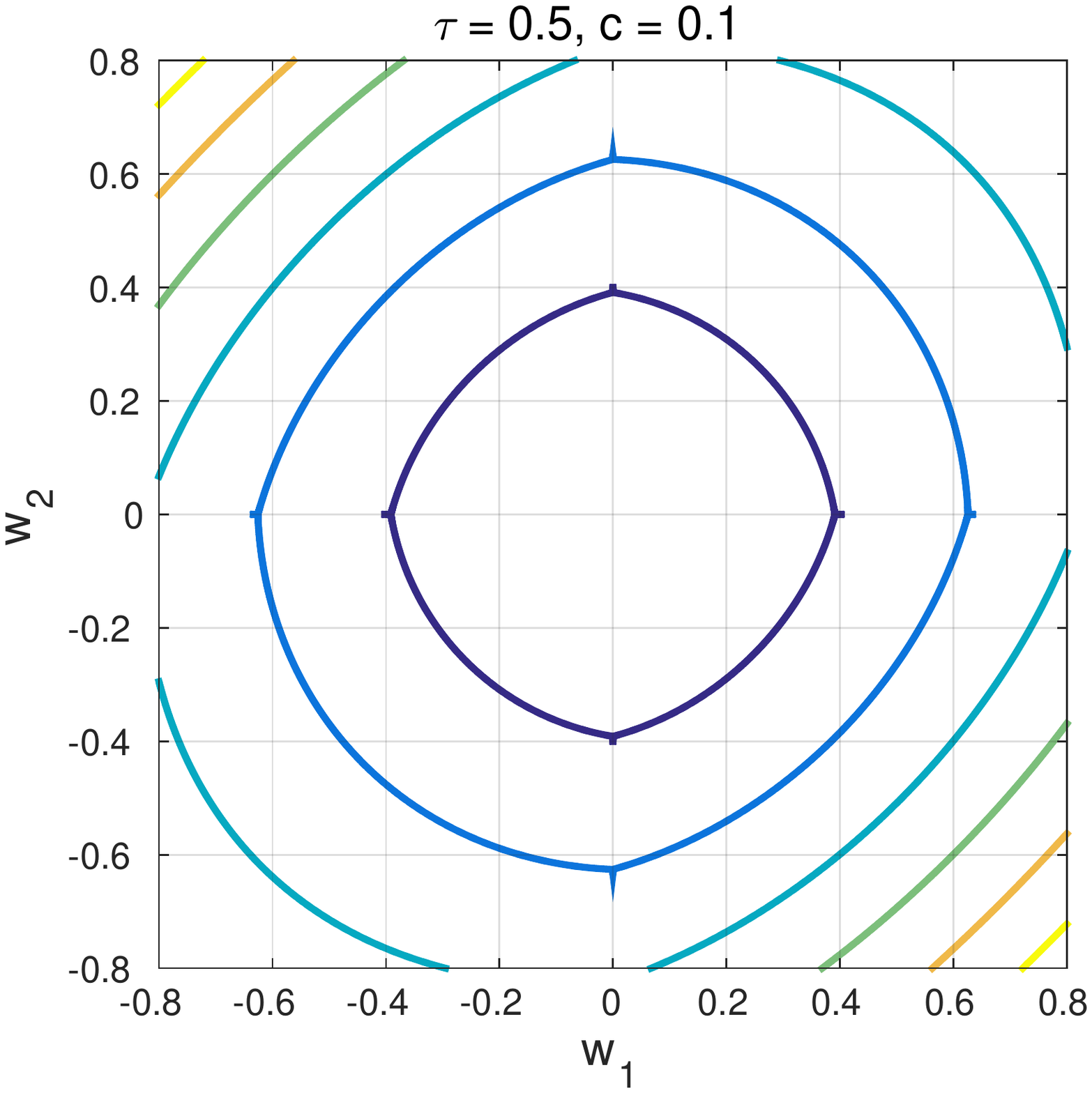}
\includegraphics[bb=70bp 180bp 495bp 642bp,scale=0.35]{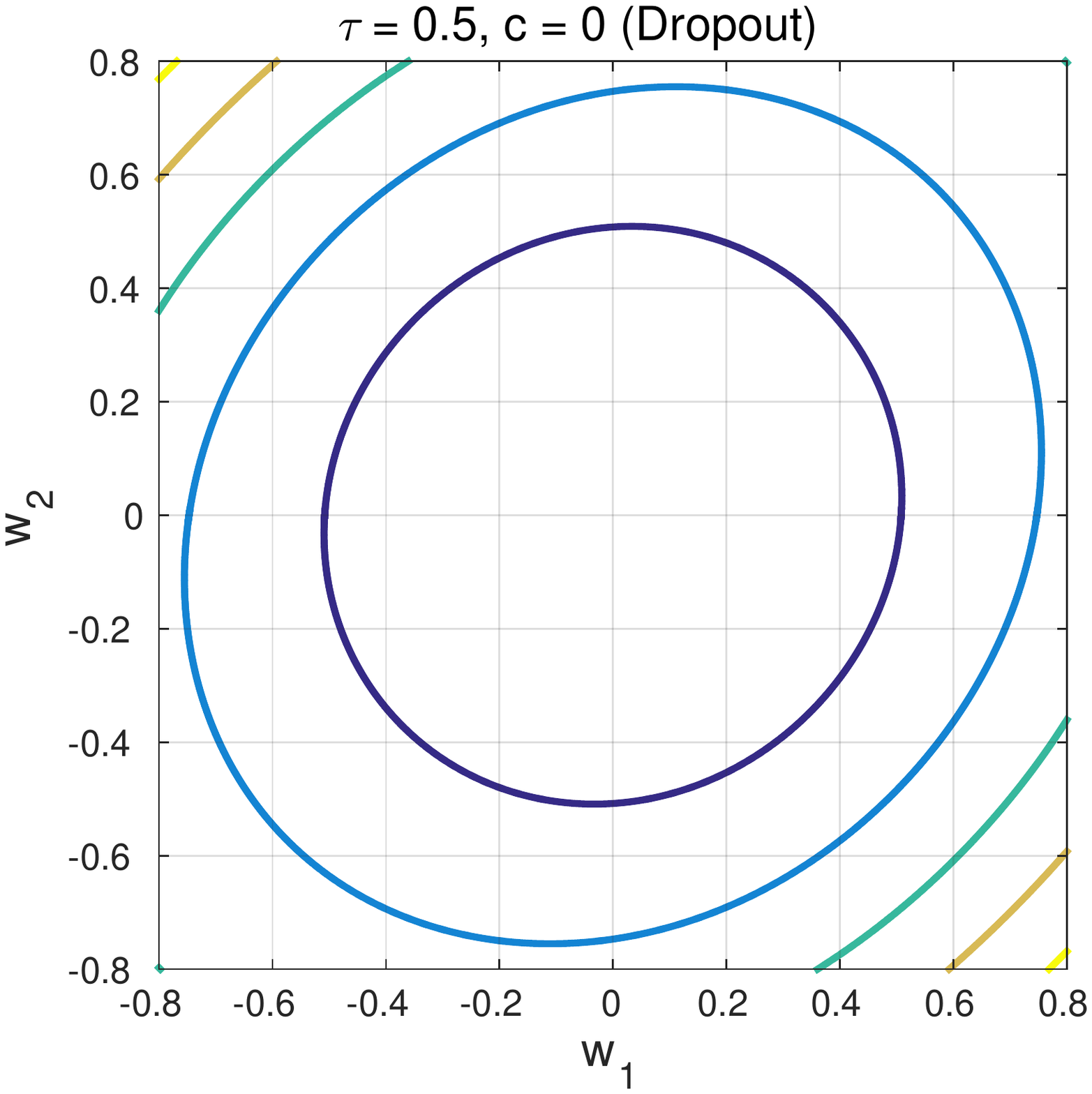}

\protect\caption{The contour plots of the regularization effect induced by Shakeout in 2D weight space with input  $\x=[1,1]^T$. Note that Dropout is a special case of Shakeout with
$c=0$.}
\label{fig:contour-of-shakeout-reg}
\end{figure*}

\textit{ii}) Logistic regression: $A(\theta)=\ln(1+\exp(\theta))$,
then
\begin{equation}
\Regw=\sum_{j=1}^{p}\ln(\frac{(1+\exp(\theta_{j-}))^{\tau}(1+\exp(\theta_{j+}))^{1-\tau}}{1+\exp(\theta)})\label{eq:sk-reg-lr}
\end{equation}
Fig. \ref{fig:contour-of-shakeout-reg} illustrates the contour of
Shakeout regularizer based on Eq. (\ref{eq:sk-reg-lr})
in the 2D weight space.
On the whole, the contour of Shakeout regularizer indicates that
the regularizer combines $L_{0}$, $L_{1}$ and $L_{2}$ regularization terms.
As $c$ goes to zero, the contour around $w=0$ becomes less sharper,
which implies hyper-parameter $c$ relates to the strength of $L_{0}$
and $L_{1}$ components. When $c=0$, Shakeout degenerates to Dropout,
the contour of which implies Dropout regularizer consists of $L_2$ regularization term.

The difference between Shakeout and Dropout regularizers is also illustrated
in Fig. \ref{fig:comparision-with-reg-approx}. We set $\tau=0.3$, $c=0.78$ for Shakeout, and $\tau=0.5$ for Dropout to make the bounds of the regularization effects of two regularizers the same. In this one dimension
circumstance, the main difference is that at $w=0$ (see the enlarged snapshot for comparison), Shakeout regularizer
is sharp and discontinuous while Dropout regularizer is smooth. Thus compared
to Dropout, Shakeout may lead to much sparser weights of the model.

To simplify the analysis and prove the intuition we have observed
in Fig. \ref{fig:contour-of-shakeout-reg} about the properties of
Shakeout regularizer, we quadratically approximate Shakeout
regularizer of Eq. (\ref{eq:shakeout-reg-accurate}) by 
\begin{equation}
\pi_{approx}(\w)=\frac{\tau}{2(1-\tau)}A^{''}(\theta)\pen
\end{equation}
The $\pen$, already shown in Eq. (\ref{eq:pen-decomp}), consists of
the combination of $L_{0}$, $L_{1}$, $L_{2}$ regularization terms. It tends to penalize
the weight whose corresponding feature's magnitude is large. Meanwhile, the weights
whose corresponding features are always zeros are less penalized. The term
$A^{''}(\theta)$ is proportional to the variance of prediction $y$ given $\x$ and $\w$.
Penalizing $A^{''}(\theta)$ encourages the weights to move towards
making the model be more "confident" about its predication, i.e. be more discriminative.

Generally speaking, Shakeout regularizer adaptively combines
$L_{0}$, $L_{1}$ and $L_{2}$ regularization terms, the property
of which matches what we have observed in Fig. \ref{fig:contour-of-shakeout-reg}.
It prefers penalizing the weights who have large magnitudes and encourages
the weights to move towards making the model more discriminative. Moreover,
the weights whose corresponding features are always zeros are less
penalized. The $L_{0}$ and $L_{1}$ components can induce sparse
weights. 

Last but not the least, we want to emphasize that when $\tau=0$, the noise is eliminated and the model becomes a standard
GLM. Moreover, Dropout can be viewed as the special case of Shakeout
when $c=0$, and a higher value of $\tau$ means a stronger $L_{2}$
regularization effect imposed on the weights. Generally, when $\tau$ is fixed ($\tau\neq0)$,
a higher value of $c$ means a stronger effect of the $L_{0}$ and
$L_{1}$ components imposed and leads to much sparser
weights of the model. We will verify this property in our experiment
section later.

\subsection{Shakeout in Multilayer Neural Networks}
It has been illustrated that Shakeout regularizes the weights in linear
modules. Linear module is the basic component of multilayer neural
networks. That is, the linear operations connect the outputs of two
successive layers. Thus Shakeout is readily applicable to the training
of multilayer neural networks. 

Considering the forward computation from layer $l$ to layer $l+1$,
for a fully-connected layer, the Shakeout forward computation is as
follows
\begin{align}
u_{i} & =\sum_{j}x_{j}[r_{j}W_{ij}+c(r_{j}-1)S_{ij}]+b_{i}\label{eq:feed-forward-1}
\end{align}
\begin{align}
x^{'}_{i} & =f(u_{i})
\end{align}
where $i$ denotes the index of the output unit of layer $l+1$, and $j$
denotes the index of the output unit of layer $l$. The output unit of a
layer is represented by $x$. The weight of the connection between
unit $x_{j}$ and unit $x^{'}_{i}$ is represented as $W_{ij}$. The bias
for the ${i}$-th unit is denoted by $b_{i}$. The $S_{ij}$ is the sign of
corresponding weight $W_{ij}$. After Shakeout operation, the linear combination
$u_{i}$ is sent to the activation function $f(\cdot)$ to obtain
the corresponding output $x^{'}_{i}$. Note that
the weights $W_{ij}$ that connect to the same input unit $x_{j}$
are controlled by the same random variable $r_{j}$.

During back-propagation, we should compute the gradients with respect
to each unit to propagate the error. In Shakeout, $\frac{\partial u_{i}}{\partial x_{j}}$
takes the form
\begin{align}
\frac{\partial u_{i}}{\partial x_{j}} & =r_{j}(W_{ij}+cS_{ij})-cS_{ij}
\end{align}
And the weights are updated following
\begin{align}
\frac{\partial u_{i}}{\partial W_{ij}} & =x_{j}(r_{j}+c(r_{j}-1)\frac{d S_{ij}}{d W_{ij}})
\end{align}
where $\frac{dS_{ij}}{dW_{ij}}$ represents the derivative
of a $\textrm{sgn}$ function. Because the $\textrm{sgn}$ function
is not continuous at zero and thus the derivative is not defined,
we approximate this derivative with $\frac{d\tanh(W_{ij})}{dW_{ij}}$.
Empirically we find that this approximation works well.

Note that the forward-backward computations with Shakeout can be easily extended to the convolutional layer.
For a convolutional layer, the Shakeout feed-forward process can be
formalized as
\begin{equation}
\mathbf{U}_{i}=\sum_{j}(\mathbf{X}_{j}\circ\mathbf{R}_{j})*\mathbf{W}_{ij}+c(\mathbf{X}_{j}\circ(\mathbf{R}_{j}-1))*\mathbf{S}_{ij}+b_{i}
\end{equation}
\begin{equation}
\mathbf{X}^{'}_{i}=f(\mathbf{U}_{i})
\end{equation}
where $\mathbf{X}_{j}$ represents the $j$-th feature map. $\mathbf{R}_{j}$
is the $j$-th random mask which has the same spatial structure (i.e.
the same height and width) as the corresponding feature map $\mathbf{X}_{j}$. $\mathbf{W}_{ij}$
denotes the kernel connecting $\mathbf{X}_{j}$ and $\mathbf{U}_{i}$.
And $\mathbf{S}_{ij}$ is set as $\mathrm{sgn}(\mathbf{W}_{ij})$. The symbol
{*} denotes the convolution operation. And the symbol $\circ$ means element-wise product.

Correspondingly, during the back-propagation process, the gradient
with respect to a unit of the layer on which Shakeout is applied takes
the form
\begin{align}
\frac{\partial\mathbf{U}_{i}(a,b)}{\partial\mathbf{X}_{j}(a-a^{'},b-b^{'})} & =\mathbf{R}_{j}(a-a^{'},b-b^{'})(\mathbf{W}_{ij}(a^{'},b^{'})+\nonumber \\
 & c\mathbf{S}_{ij}(a^{'},b^{'}))-c\mathbf{S}_{ij}(a^{'},b^{'})
\end{align}
where $(a,b)$ means the position of a unit in the output feature map of a layer, and
$(a^{'},b^{'})$ represents the position of a weight in the corresponding kernel.

The weights are updated following
\begin{multline}
\frac{\partial\mathbf{U}_{i}(a,b)}{\partial\mathbf{W}_{ij}(a^{'},b^{'})}=\mathbf{X}_{j}(a-a^{'},b-b^{'})(\mathbf{R}_{j}(a-a^{'},b-b^{'})\\
+c(\mathbf{R}_{j}(a-a^{'},b-b^{'})-1)\frac{d \mathbf{S}_{ij}(a^{'},b^{'})}{d \mathbf{W}_{ij}(a^{'},b^{'})})
\end{multline}


\section{Experiments}
In this section, we report empirical evaluations of Shakeout in training
deep neural networks on representative datasets. 
The experiments are performed on three kinds of
image datasets: the hand-written image dataset MNIST  \cite{lecun1998gradient}, the CIFAR-10 image dataset \cite{krizhevsky2009learning} and
the ImageNet-2012 dataset \cite{ILSVRC15}. MNIST consists of 60,000+10,000 (training+test)
28$\times$28 images of hand-written digits. CIFAR-10 contains 50,000+10,000
(training+test) 32$\times$32 images of 10 object classes. ImageNet-2012
consists of 1,281,167+50,000+150,000 (training+validation+test)
variable-resolution images of 1000 object classes.
We first demonstrate
that Shakeout leads to sparse models as our theoretical analysis implies
under the unsupervised setting. Then we show that for the classification task, the sparse models have desirable generalization performances. Further, we illustrate the regularization effect 
of Shakeout on the weights in the classification task. 
Moreover, the effect of Shakeout on stabilizing the training processes of the deep architectures is demonstrated.
Finally, we give some
practical recommendations of Shakeout. All the experiments are implemented
based on the modifications of \textit{Caffe} library \cite{jia2014caffe}.
Our code is released on the github: https://github.com/kgl-prml/shakeout-for-caffe.

\subsection{\label{sub:autoencoder-weight-sparsity}Shakeout  and Weight
Sparsity}
Since Shakeout implicitly imposes $L_{0}$ penalty and $L_{1}$ penalty
of the weights, we expect the weights of neural networks learned by
Shakeout contain more zeros than those learned by the standard back-propagation (BP)
\cite{williams1986learning} or Dropout \cite{hinton2012improving}.
In this experiment, we employ an autoencoder model for the MNIST
hand-written data, train the model using
standard BP, Dropout and Shakeout, respectively, and compare the degree
of sparsity of the weights of the learned encoders. For the purpose
of demonstration, we employ the simple autoencoder with one hidden
layer of 256 units; Dropout and Shakeout are applied on the input pixels.

\begin{figure}
\centering
\includegraphics[bb=0bp 180bp 595bp 662bp,scale=0.4]{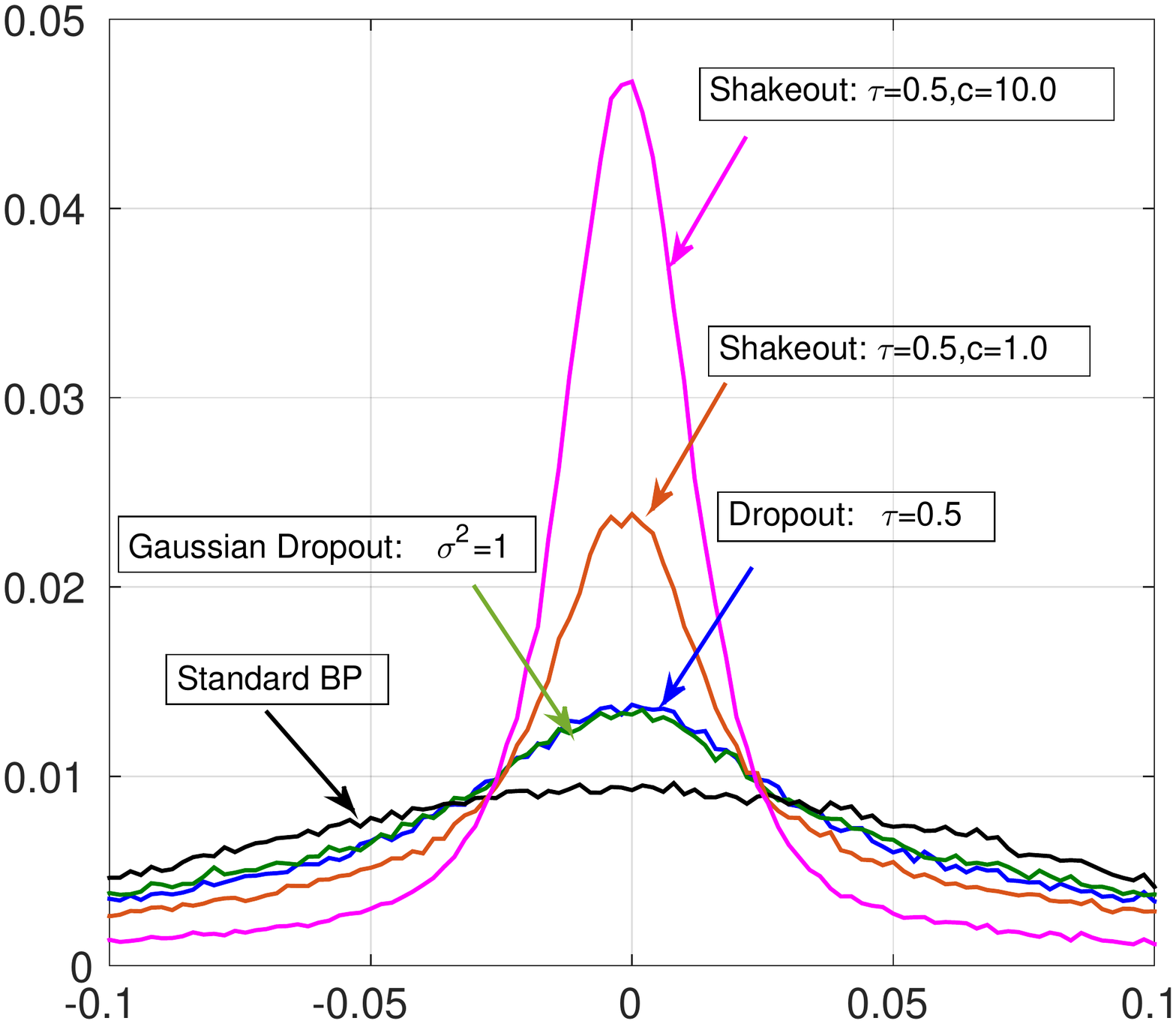}
\protect\caption{Distributions
of the weights of the autoencoder models learned by different training
approaches. Each curve in the figure shows the frequencies of the weights
of an autoencoder taking particular values, i.e. the
empirical population densities of the weights. The five curves correspond
to five autoencoders learned by standard back-propagation, Dropout ($\tau=0.5$), 
Gaussian Dropout ($\sigma^{2}=1$) and Shakeout ($\tau=0.5$, $c=\{1,10\}$). The sparsity
of the weights obtained via Shakeout can be seen by comparing the
curves. }
\label{fig:The-distributions-of-AE}
\end{figure}

\begin{figure*}[!t]
\centering
\subfloat[standard BP]{\includegraphics[scale=0.38]{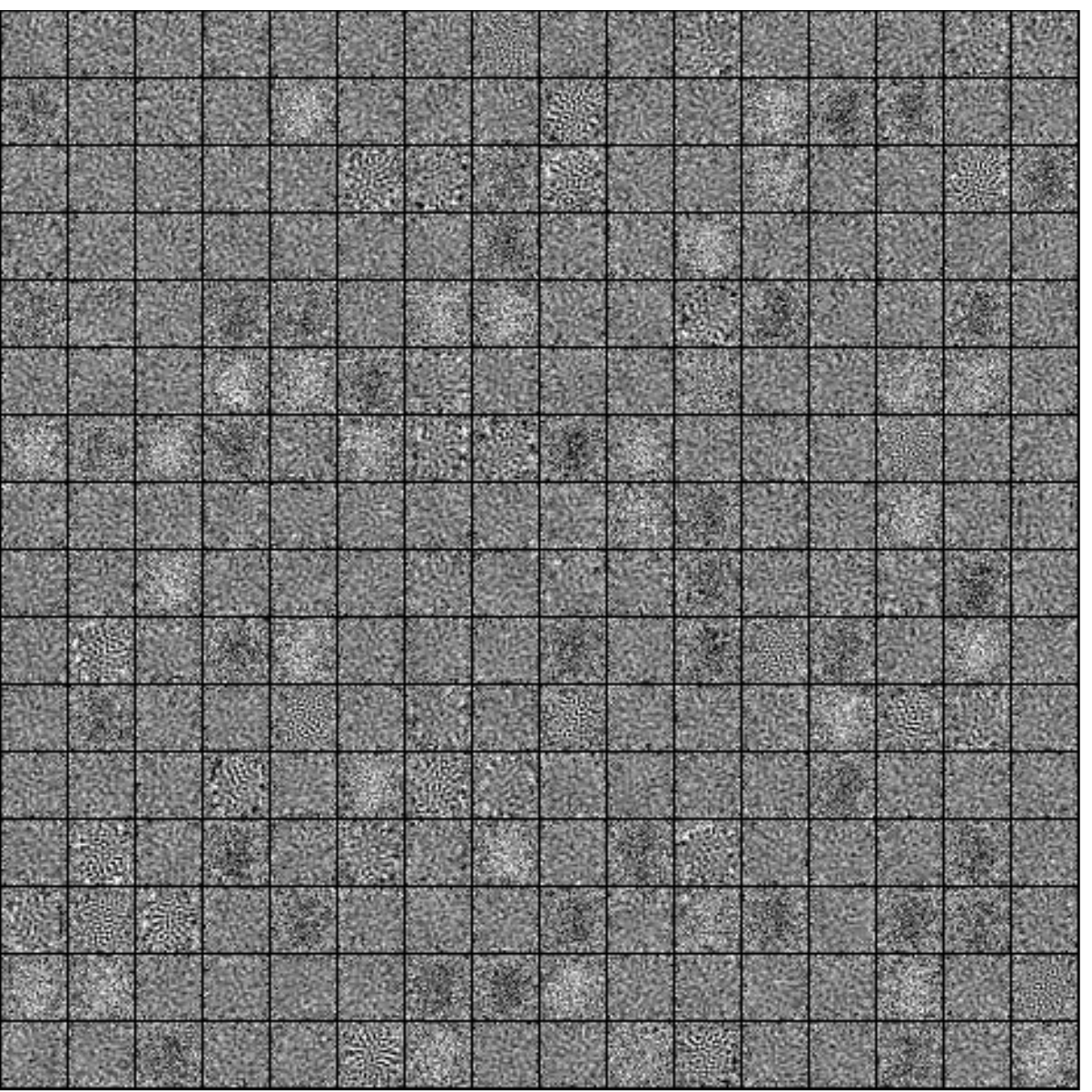}} 
\hfil
\subfloat[Dropout: $\tau=0.5$]{\includegraphics[scale=0.38]{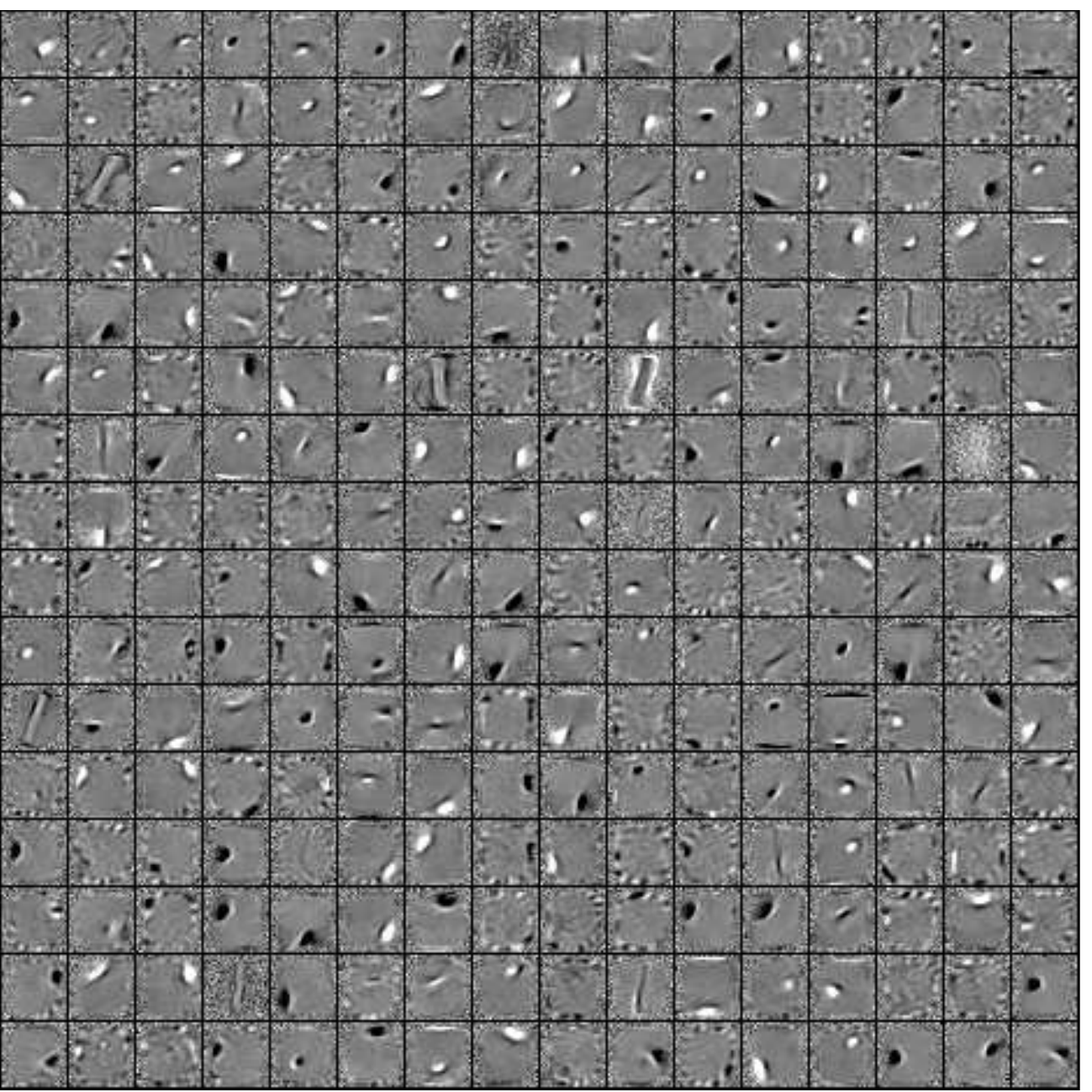}}
\hfil
\subfloat[Shakeout: $\tau=0.5$, $c=0.5$]{\includegraphics[scale=0.38]{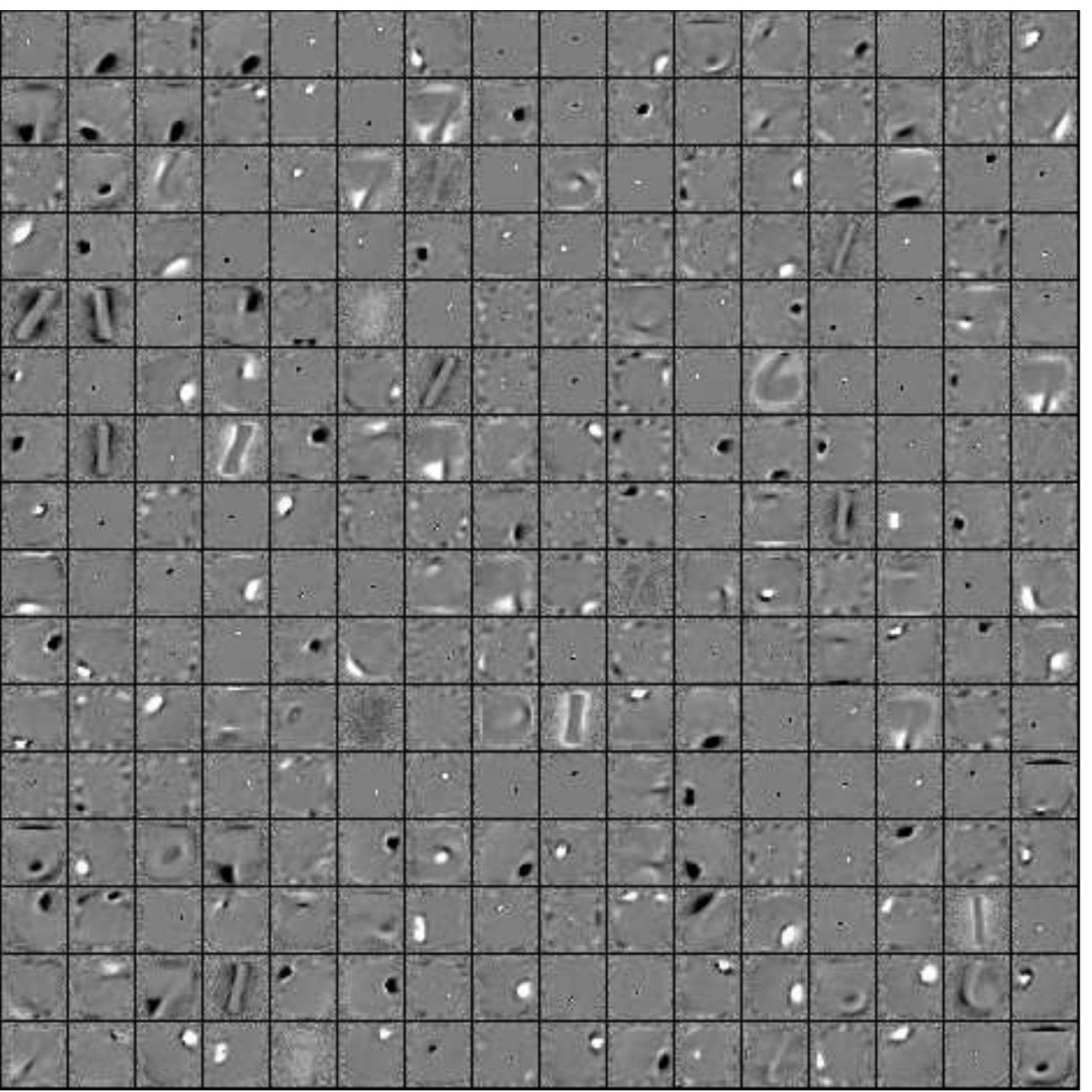}}

\caption{Features captured
by the hidden units of the autoencoder models
learned by different training methods. The features
captured by a hidden unit are represented by a group of weights that
connect the image pixels with this corresponding hidden unit. One
image patch in a sub-graph corresponds to the features captured by
one hidden unit.}
\label{fig:Learned-weights-of-vi}
\end{figure*}

To verify the regularization effect, we compare the weights of the
four autoencoders trained under different settings which correspond
to standard BP, Dropout ($\tau=0.5$) and Shakeout ($\tau=0.5$, $c=\{1,10\}$). All the training methods aim to produce hidden units
which can capture good visual features of the handwritten digits.
The statistical traits of these different resulting weights are shown
in Fig. \ref{fig:The-distributions-of-AE}. Moreover, Fig. \ref{fig:Learned-weights-of-vi}
shows the features captured by each hidden unit of the autoencoders.

As shown in the Fig. \ref{fig:The-distributions-of-AE}, the probability
density of weights around the zero obtained by standard BP training
is quite small compared to the one obtained either by Dropout or Shakeout.
This indicates the strong regularization effect induced by Dropout
and Shakeout. Furthermore, the sparsity level of weights obtained
from training by Shakeout is much higher than the one obtained from
training by Dropout. Using the same $\tau$, increasing $c$ makes
the weights much sparser, which is consistent with the characteristics
of $L_{0}$ penalty and $L_{1}$ penalty induced by Shakeout. Intuitively,
we can find that due to the induced $L_{2}$ regularization, the distribution
of weights obtained from training by the Dropout is like a Gaussian,
while the one obtained from training by Shakeout is more like a Laplacian
because of the additionally induced $L_{1}$ regularization. Fig. \ref{fig:Learned-weights-of-vi}
shows that features captured by the hidden units via standard BP training
are not directly interpretable, corresponding to insignificant variants
in the training data. Both Dropout and Shakeout suppress
irrelevant weights by their regularization effects, where Shakeout
produces much sparser and more global features thanks to the combination
of $L_{0}$, $L_{1}$ and $L_{2}$ regularization terms.

The autoencoder trained by Dropout or Shakeout can be viewed as the denosing autoencoder, where Dropout or Shakeout injects special kind of noise into the inputs. Under this unsupervised setting, the denoising criterion (i.e. minimizing the error between imaginary images reconstructed from the noisy inputs and the real images without noise) is to guide the learning of useful high level feature representations \cite{vincent2008extracting,vincent2010stacked}.
To verify that Shakeout helps learn better feature representations, we adopt the hidden layer activations 
as features to train SVM classifiers, and the classification accuracies on test set 
for standard BP, Dropout and Shakeout are 95.34\%, 96.41\% and 96.48\%,
respectively. We can see that Shakeout leads to much sparser
weights without defeating the main objective.

Gaussian Dropout has similar effect on the model training as standard Dropout \cite{srivastava2014dropout}, which multiplies the activation
of each unit by a Gaussian variable with mean 1 and variance 
$\sigma^{2}$. The relationship between $\sigma^{2}$ and $\tau$ is that
$\sigma^{2}=\frac{\tau}{1-\tau}$. The distribution of the weights
trained by Gaussian Dropout ($\sigma^{2}=1$, i.e. $\tau=0.5$) 
is illustrated in Fig. \ref{fig:The-distributions-of-AE}.
From Fig. \ref{fig:The-distributions-of-AE}, we find no notable 
statistical difference between
two kinds of Dropout implementations which all exhibit a kind of
$L_{2}$ regularization effect on the weights. The classification
performances of SVM classifiers on test set 
based on the hidden layer activations as extracted features
for both kinds of Dropout implementations are quite
similar (i.e. $96.41\%$ and $96.43\%$ for standard and Gaussian
Dropout respectively). Due to these observations, we conduct the
following classification experiments using standard Dropout as a representative
implementation (of Dropout) for comparison.


\subsection{Classification Experiments}
Sparse models often indicate lower complexity and better generalization performance
\cite{tibshirani1996regression,zou2005regularization,olshausen1997sparse,yuan2013efficient}.
To verify the effect of $L_{0}$ and $L_{1}$ regularization terms induced by Shakeout on the model performance, we apply
Shakeout, along with Dropout and standard BP, on training representative deep neural
networks for classification tasks. In all of our classification experiments,
the hyper-parameters $\tau$ and $c$ in Shakeout, and the hyper-parameter $\tau$ in Dropout
are determined by validation.
\subsubsection{MNIST}
\begin{figure*}
\centering
\subfloat[Fully-connected neural network]{\includegraphics[bb=0bp 180bp 595bp 600bp,scale=0.35]{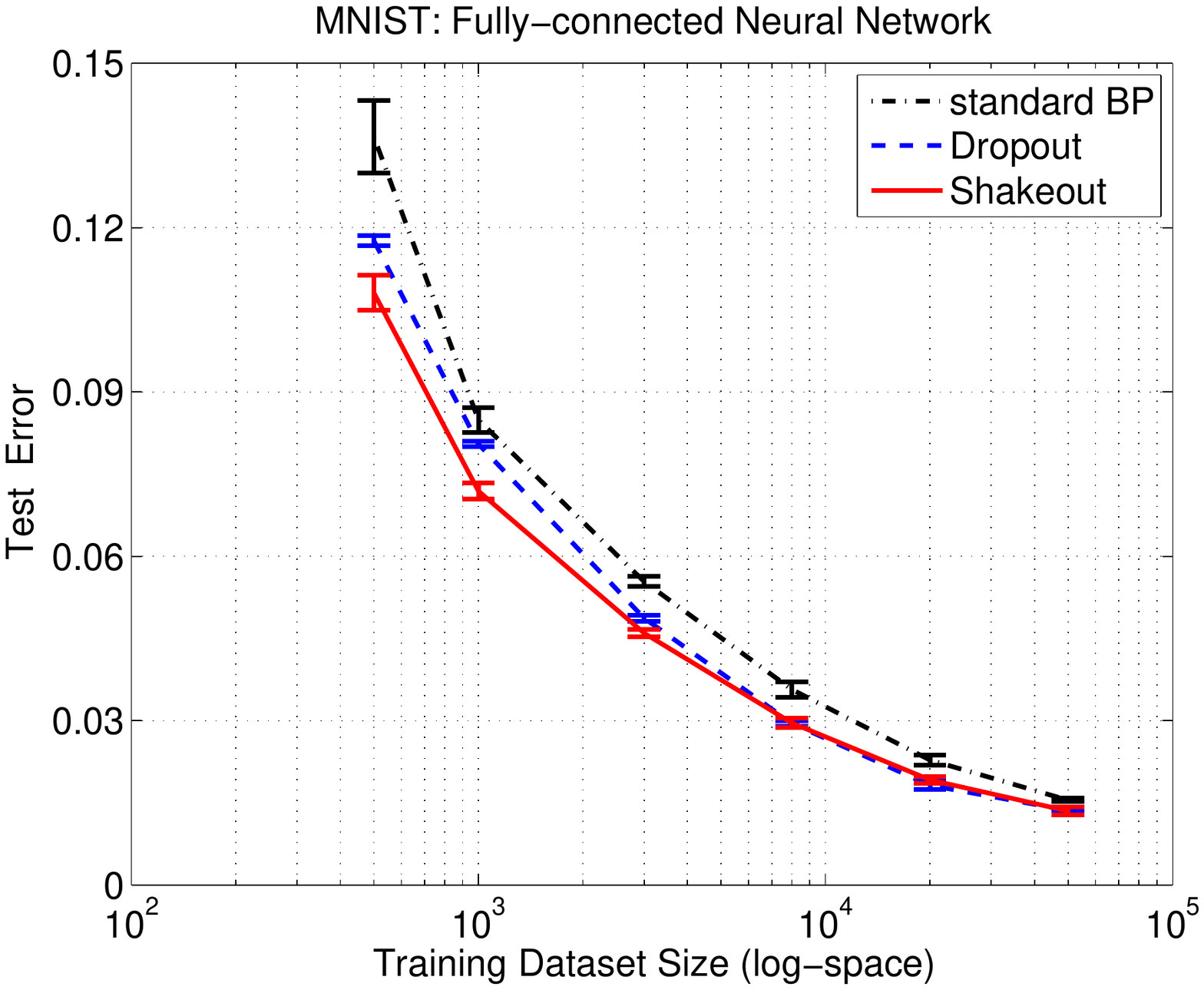}}
\hfil
\subfloat[Convolutional neural network]{\includegraphics[bb=0bp 180bp 595bp 600bp,scale=0.35]{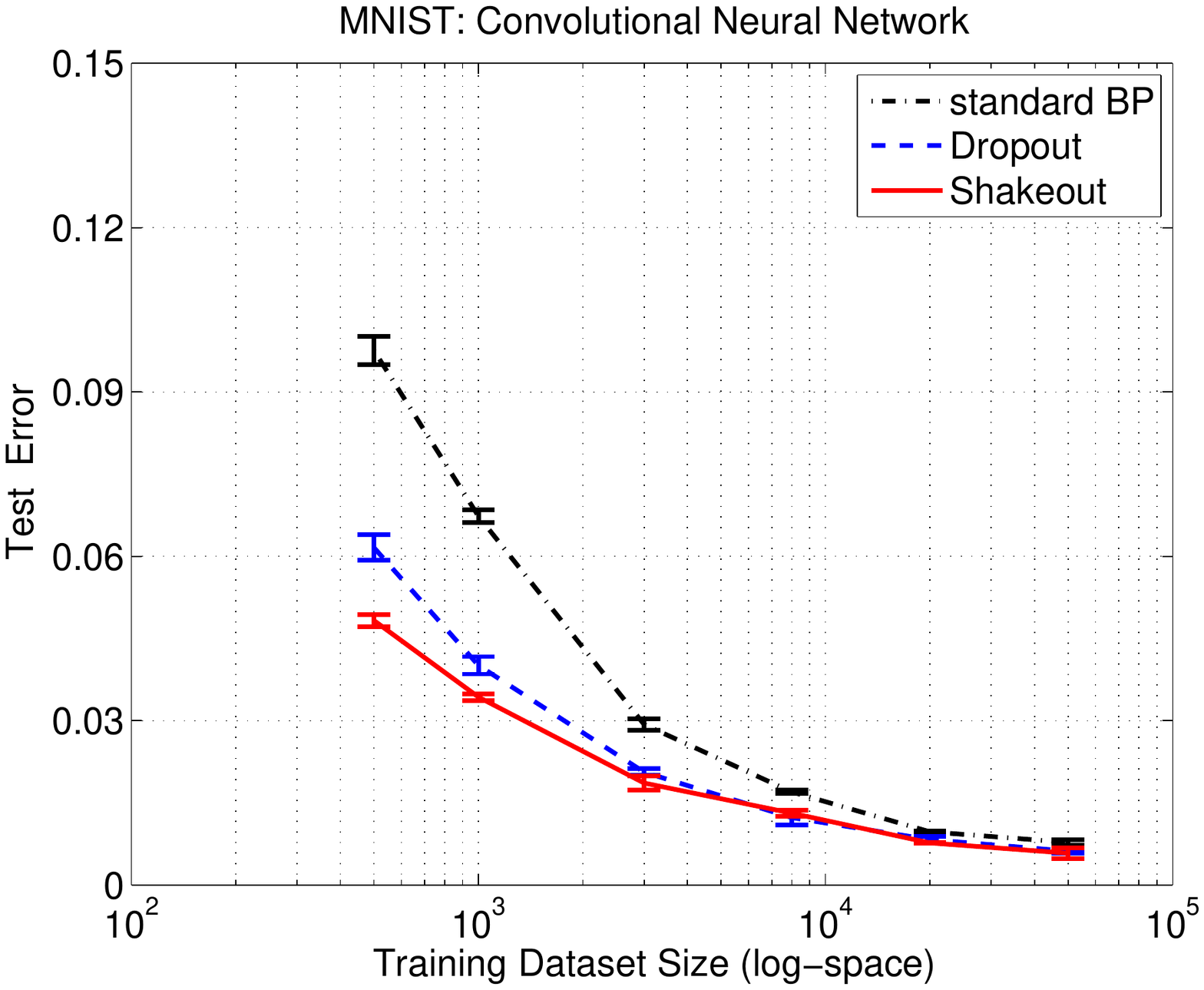}}

\protect\caption{Classification of two kinds
of neural networks on MNIST using training sets of different sizes.
The curves show the performances of the models trained by standard
BP, and those by Dropout and Shakeout applied on
the hidden units of the fully-connected layer.}
\label{fig:MNIST-test-classification}
\end{figure*}

We train two different neural networks, a shallow fully-connected
one and a deep convolutional one. For the fully-connected neural network,
a big hidden layer size is adopted with its value at 4096. The non-linear
activation unit adopted is the rectifier linear unit (ReLU). The deep
convolutional neural network employed is based on the modifications
of the LeNet \cite{lecun1998gradient}, which contains two convolutional
layers and two fully-connected layers. The detailed architecture information
of this convolutional neural network is described in Tab. \ref{tab:The-architechture-of-conv}.
\begin{table}
\protect\caption{\label{tab:The-architechture-of-conv}The architecture of convolutional
neural network adopted for MNIST classification experiment }

\centering{}%
\begin{tabular}{|c|c|c|c|c|}
\hline 
Layer & 1 & 2 & 3 & 4\tabularnewline
\hline 
\hline 
Type & conv. & conv. & FC & FC\tabularnewline
\hline 
Channels & 20 & 50 & 500 & 10\tabularnewline
\hline 
Filter size & $5\times5$ & $5\times5$ & - & -\tabularnewline
\hline 
Conv. stride & 1 & 1 & - & -\tabularnewline
\hline 
Pooling type & max & max & - & -\tabularnewline
\hline 
Pooling size & $2\times2$ & $2\times2$ & - & -\tabularnewline
\hline 
Pooling stride & 2 & 2 & - & -\tabularnewline
\hline 
Non-linear & ReLU & ReLU & ReLU & Softmax\tabularnewline
\hline 
\end{tabular}
\end{table}
We separate 10,000 training samples from original training dataset
for validation. The results are shown in Tab. \ref{tab:MNIST-test-classification-fc}
and Tab. \ref{tab:MNIST-test-classification-cov}. Dropout and Shakeout
are applied on the hidden units of the fully-connected
layer. The table compares the errors of the networks trained by standard
back-propagation, Dropout and Shakeout. The mean and standard deviation
of the classification errors are obtained by 5 runs of the experiment
and are shown in percentage. We can see from the results that when
the training data is not sufficient enough, due to over-fitting, all
the models perform worse. However, the models trained by Dropout and
Shakeout consistently perform better than the one trained by standard
BP. Moreover, when the training data is scarce, Shakeout leads to
superior model performance compared to the Dropout. Fig. \ref{fig:MNIST-test-classification}
shows the results in a more intuitive way.
\begin{table}[!t]
\protect\caption{Classification on MNIST using
training sets of different sizes: fully-connected neural network}
\label{tab:MNIST-test-classification-fc}
\centering{}%
\begin{tabular}{|c|c|c|c|}
\hline 
Size & std-BP & Dropout & Shakeout \tabularnewline
\hline 
\hline 
500 & 13.66$\pm$0.66 & 11.76$\pm$0.09 & \textbf{10.81}$\pm$0.32\tabularnewline
\hline 
1000 & 8.49$\pm$0.23 & 8.05$\pm$0.05 & \textbf{7.19}$\pm$0.15\tabularnewline
\hline 
3000 & 5.54$\pm$0.09 & 4.87$\pm$0.06 & \textbf{4.60}$\pm$0.07\tabularnewline
\hline 
8000 & 3.57$\pm$0.14 & \textbf{2.95}$\pm$0.05 & 2.96$\pm$0.09\tabularnewline
\hline 
20000 & 2.28$\pm$0.09 & \textbf{1.82}$\pm$0.07 & 1.92$\pm$0.06\tabularnewline
\hline 
50000 & 1.55$\pm$0.03 & 1.36$\pm$0.03 & \textbf{1.35}$\pm$0.07\tabularnewline
\hline 
\end{tabular}
\end{table}

\subsubsection{CIFAR-10}\label{sec:cifar-10-sec}
\begin{figure}
\centering
\includegraphics[bb=0bp 200bp 595bp 620bp,scale=0.35]{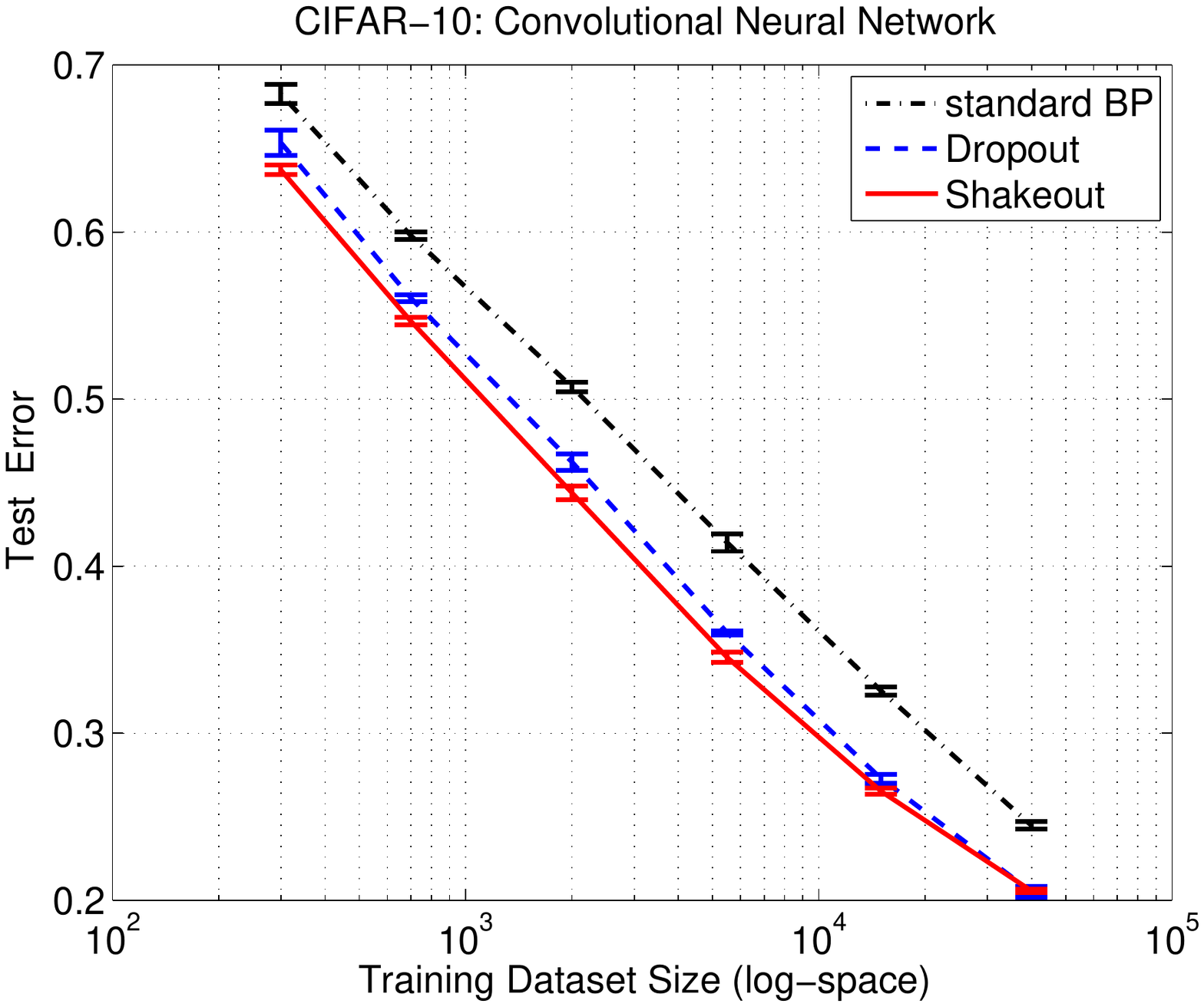}

\protect\caption{Classification on CIFAR-10
using training sets of different sizes. The curves show the performances
of the models trained by standard BP, and those by Dropout and Shakeout
applied on the hidden units of the fully-connected
layer.}
\label{fig:CIFAR10-test-classification-quick}
\end{figure}

We use the simple convolutional network feature extractor described
in cuda-convnet (layers-80sec.cfg) \cite{krizhevskycuda}. We apply
Dropout and Shakeout on the first fully-connected layer.
We call this architecture ``AlexFastNet'' for the convenience of
description. In this experiment, 10,000 colour images are separated
from the training dataset for validation and no data augmentation
is utilized. The per-pixel mean computed over the training set is
subtracted from each image. We first train for 100 epochs with an
initial learning rate of 0.001 and then another 50 epochs with the
learning rate of 0.0001. The mean and standard deviation of the classification
errors are obtained by 5 runs of the experiment and are shown in percentage.
\begin{table}[!t]
\protect\caption{Classification on MNIST
using training sets of different sizes: convolutional neural network}
\label{tab:MNIST-test-classification-cov}

\centering{}%
\begin{tabular}{|c|c|c|c|}
\hline 
Size & std-BP & Dropout & Shakeout\tabularnewline
\hline 
\hline 
500 & 9.76$\pm$0.26 & 6.16$\pm$0.23 & \textbf{4.83}$\pm$0.11\tabularnewline
\hline 
1000 & 6.73$\pm$0.12 & 4.01$\pm$0.16 & \textbf{3.43}$\pm$0.06\tabularnewline
\hline 
3000 & 2.93$\pm$0.10 & 2.06$\pm$0.06 & \textbf{1.86}$\pm$0.13\tabularnewline
\hline 
8000 & 1.70$\pm$0.03 & \textbf{1.23}$\pm$0.13 & 1.31$\pm$0.06\tabularnewline
\hline 
20000 & 0.97$\pm$0.01 & 0.83$\pm$0.06 & \textbf{0.77}$\pm$$0$.001\tabularnewline
\hline 
50000 & 0.78$\pm$0.05 & 0.62$\pm$0.04 & \textbf{0.58}$\pm$0.10\tabularnewline
\hline 
\end{tabular}
\end{table}
\begin{table}[!t]
\protect\caption{Classification on CIFAR-10
using training sets of different sizes: AlexFastNet }
\label{tab:CIFAR10-test-classification-quick}

\centering{}%
\begin{tabular}{|c|c|c|c|}
\hline 
Size & std-BP & Dropout & Shakeout\tabularnewline
\hline 
\hline 
300 & 68.26$\pm$0.57 & 65.34$\pm$0.75 & \textbf{63.71}$\pm$0.28\tabularnewline
\hline 
700 & 59.78$\pm$0.24 & 56.04$\pm$0.22 & \textbf{54.66}$\pm$0.22\tabularnewline
\hline 
2000 & 50.73$\pm$0.29 & 46.24$\pm$0.49 & \textbf{44.39}$\pm$0.41\tabularnewline
\hline 
5500 & 41.41$\pm$0.52 & 36.01$\pm$0.13 & \textbf{34.54}$\pm$0.31\tabularnewline
\hline 
15000 & 32.53$\pm$0.25 & 27.28$\pm$0.26 & \textbf{26.53}$\pm$0.17\tabularnewline
\hline 
40000 & 24.48$\pm$0.23 & \textbf{20.50}$\pm$0.32 & 20.56$\pm$0.12\tabularnewline
\hline 
\end{tabular}
\end{table}
As shown in Tab. \ref{tab:CIFAR10-test-classification-quick}, the
performances of models trained by Dropout and Shakeout are consistently
superior to the one trained by standard BP. Furthermore, the model
trained by Shakeout also outperforms the one trained by Dropout when
the training data is scarce. Fig. \ref{fig:CIFAR10-test-classification-quick}
shows the results in a more intuitive way.

To test the performance of Shakeout on a much deeper architecture,
we also conduct experiments based on the Wide Residual
Network (WRN) \cite{DBLP:conf/bmvc/ZagoruykoK16}. The configuration
of WRN adopted is WRN-16-4, which means WRN has 16 
layers in total and the number of feature maps for 
the convolutional layer of each residual block is 4 times as the corresponding original one
\cite{he2016identity}. Because the complexity is much higher than
that of ``AlexFastNet'', the experiments are performed on relatively
larger training sets with sizes of 15000, 40000, 50000. Dropout and
Shakeout are applied on the second convolutional layer of each residual
block, following the protocol in \cite{DBLP:conf/bmvc/ZagoruykoK16}. 
All the training starts from the same initial weights. 
Batch Normalization is applied the same way as \cite{DBLP:conf/bmvc/ZagoruykoK16} 
to promote the optimization.
No data-augmentation or data pre-processing is adopted. All the other hyper-parameters
other than $\tau$ and $c$ are set the same as \cite{DBLP:conf/bmvc/ZagoruykoK16}.
The results are listed in Tab. \ref{tab:w16-4-cifar10}. For the training
of CIFAR-10 with 50000 training samples, we adopt the same hyper-parameters
as those chosen in the training with training set size at 40000. From
Tab. \ref{tab:w16-4-cifar10}, we can arrive at the same conclusion
as previous experiments, i.e. the performances of the models trained
by Dropout and Shakeout are consistently superior to the one trained
by standard BP. Moreover, Shakeout outperforms Dropout when the data is scarce. 
\begin{table}[!t]
\begin{centering}
\protect\caption{Classification on CIFAR-10 using training
sets of different sizes: WRN-16-4 }
\label{tab:w16-4-cifar10}

\par\end{centering}

\centering{}%
\begin{tabular}{|c|c|c|c|}
\hline 
Size & std-BP & Dropout & Shakeout\tabularnewline
\hline 
\hline 
15000 & 20.95 & 15.05 & \textbf{14.68}\tabularnewline
\hline 
40000 & 15.37 & 9.32 & \textbf{9.01}\tabularnewline
\hline 
50000 & 14.39 & 8.03 & \textbf{7.97}\tabularnewline
\hline 
\end{tabular}
\end{table}

\begin{figure*}[!t]
\centering
\subfloat[AlexNet
FC7 layer]{\includegraphics[bb=0bp 180bp 595bp 650bp,scale=0.35]{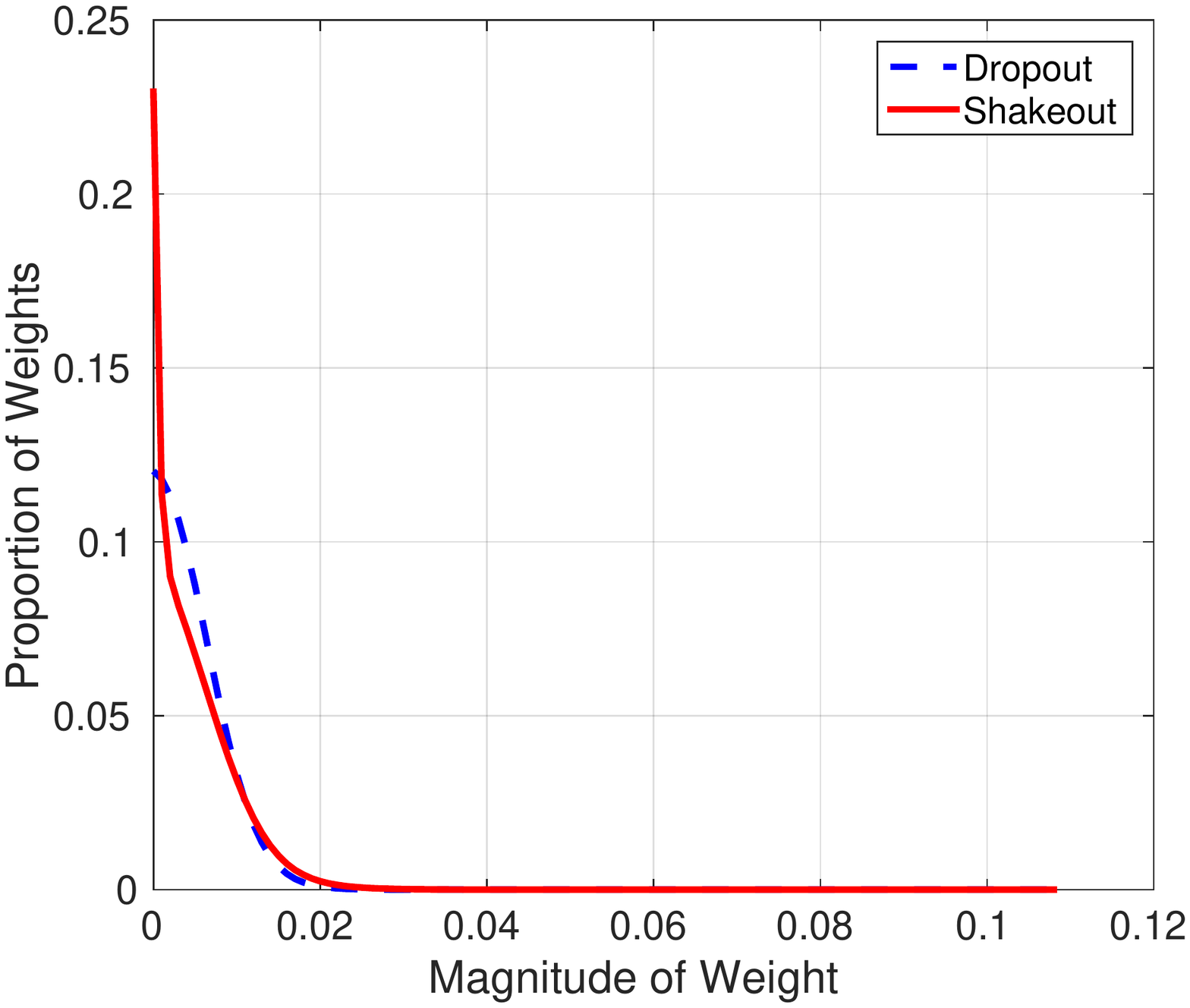}}
\hfil
\subfloat[AlexNet
FC8 layer]{\includegraphics[bb=30bp 180bp 595bp 650bp,scale=0.35]{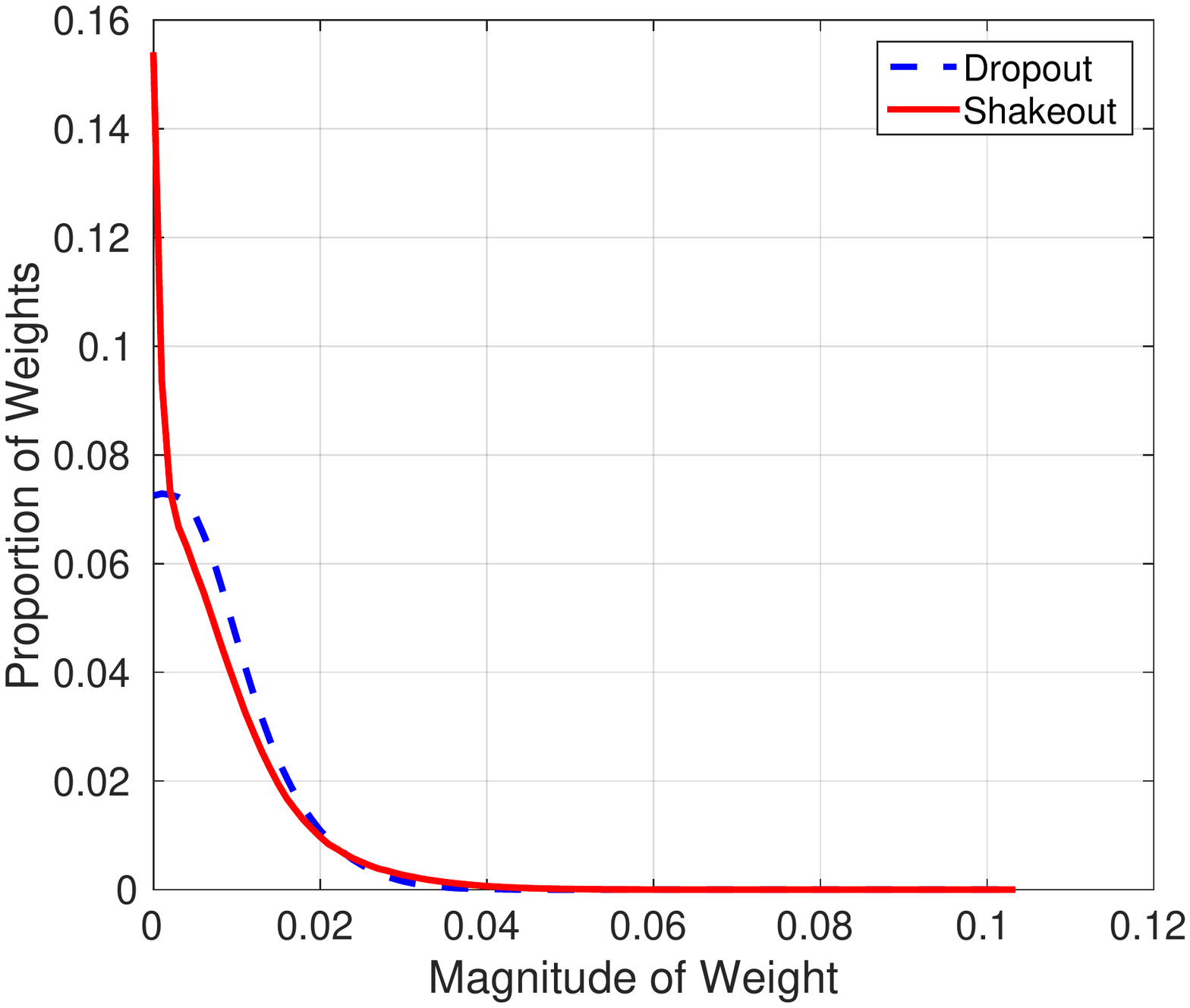}}

\protect\caption{Comparison of the distributions of
the magnitude of weights trained by Dropout and Shakeout. The experiments
are conducted using AlexNet on ImageNet-2012 dataset. Shakeout
or Dropout is applied on the last two fully-connected layers, i.e.
FC7 layer and FC8 layer.}
\label{fig:classification-sparse}
\end{figure*}
\begin{figure*}[!t]
\centering
\subfloat[AlexNet
FC7 layer]{\includegraphics[bb=0bp 180bp 595bp 650bp,scale=0.35]{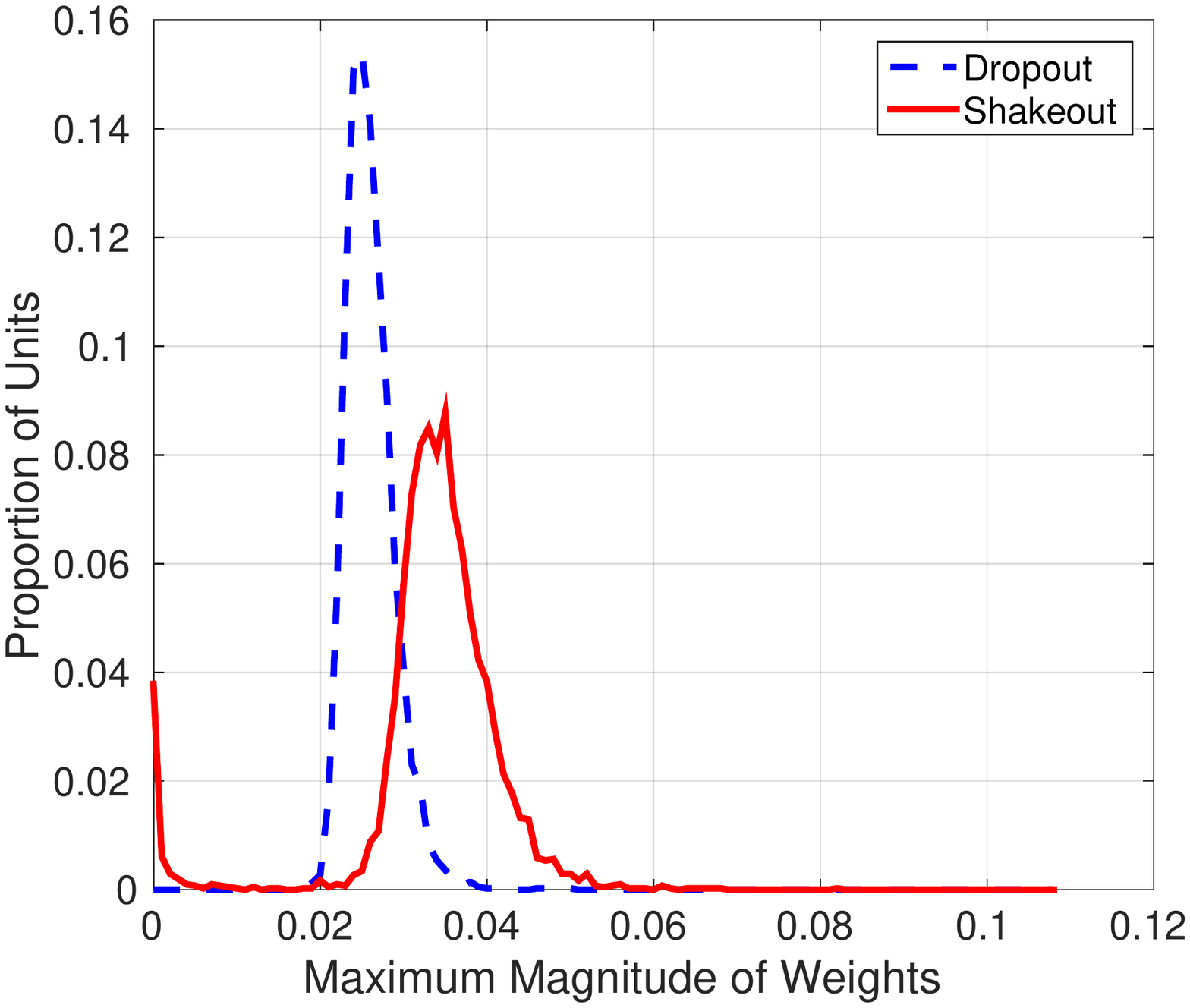}}
\hfil
\subfloat[AlexNet
FC8 layer]{\includegraphics[bb=30bp 180bp 595bp 650bp,scale=0.35]{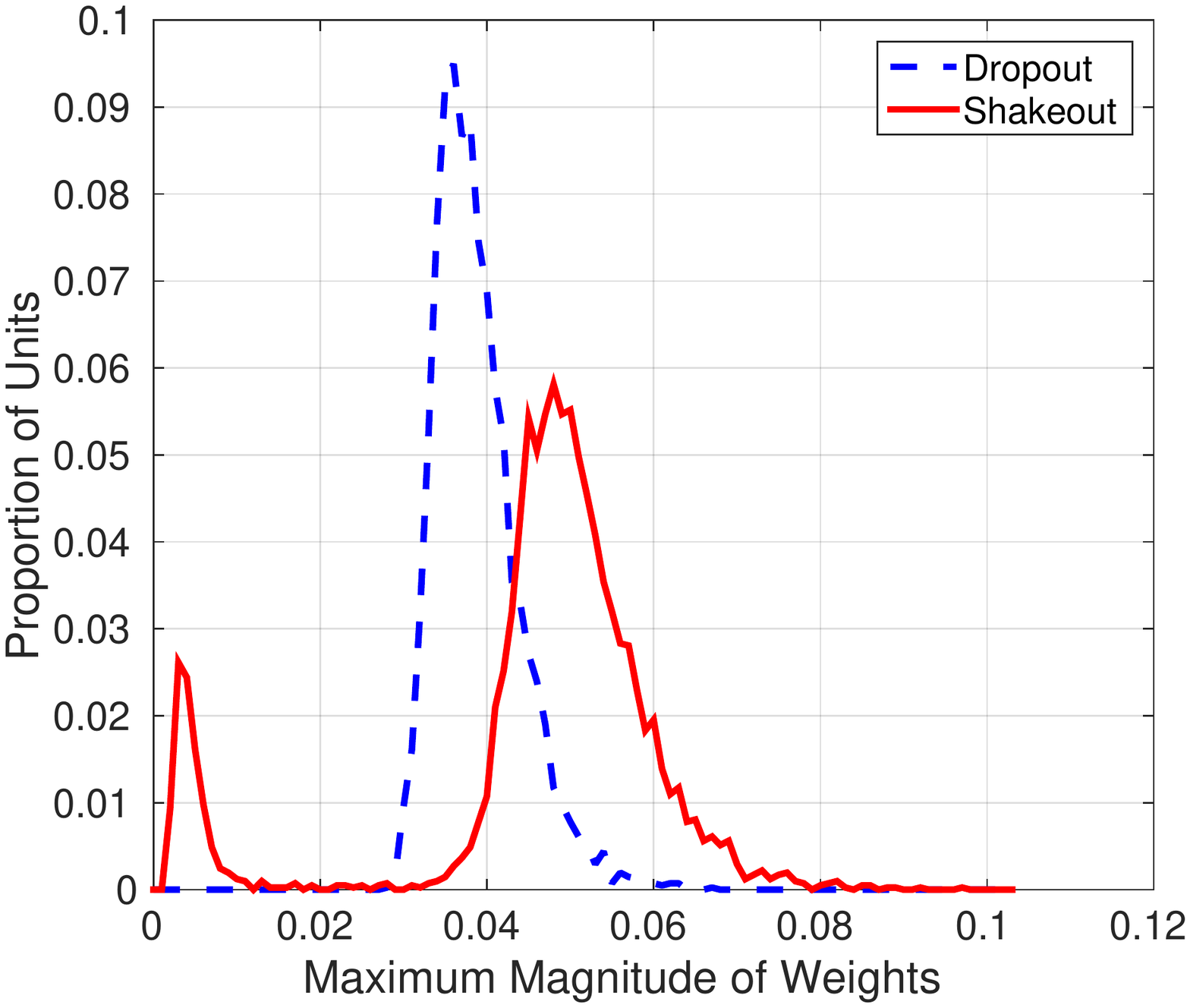}}

\caption{Distributions of the maximum magnitude
of the weights connected to the same input unit of a layer. The maximum magnitude
of the weights connected to one input unit can be regarded as a metric
of the importance of that unit. The experiments
are conducted using AlexNet on ImageNet-2012 dataset. For Shakeout, the units
can be approximately separated into two groups and the one around
zero is less important than the other, whereas for Dropout, the
units are more concentrated.}
\label{fig:grouping-effect}
\end{figure*}

\subsubsection{Regularization Effect on the Weights}
Shakeout is a different way to regularize the training process
of deep neural networks from Dropout. For a GLM model, we have proved
that the regularizer induced by Shakeout adaptively combines $L_{0}$,
$L_{1}$ and $L_{1}$ regularization terms. In section \ref{sub:autoencoder-weight-sparsity},
we have demonstrated that for a one-hidden layer autoencoder, it
leads to much sparser weights of the model. In this section, we will
illustrate the regularization effect of Shakeout on the weights 
in the classification task and make a comparison to that of Dropout.

The results shown in this section are mainly based on the experiments conducted on ImageNet-2012 dataset using the representative
deep architecture: AlexNet \cite{krizhevsky2012imagenet}. 
For AlexNet, we apply Dropout or Shakeout
on layers FC7 and FC8 which are the last two fully-connected layers.
We train the model from the scratch and obtain the comparable classification 
performances on validation set for Shakeout (top-1 error: 42.88\%; top-5 error: 19.85\%)
and Dropout (top-1 error: 42.99\%; top-5 error: 19.60\%). The model
is trained based on the same hyper-parameter settings provided by
Shelhamer in \textit{Caffe} \cite{jia2014caffe} other than the hyper-parameters
$\tau$ and $c$ for Shakeout. The initial weights for training by Dropout
and Shakeout are kept the same.

Fig. \ref{fig:classification-sparse} illustrates the distributions
of the magnitude of weight resulted by Shakeout and Dropout. It can be seen that
the weights learned by Shakeout are much sparser than those learned
by Dropout, due to the implicitly induced $L_{0}$ and $L_{1}$ components. 

The regularizer induced by Shakeout not only contains $L_{0}$ and
$L_{1}$ regularization terms but also contains $L_{2}$ regularization term, the combination of
which is expected to discard a group of weights simultaneously. In
Fig. \ref{fig:grouping-effect}, we use the maximum magnitude of the
weights connected to one input unit of a layer to represent the
importance of that unit for the subsequent output units. From Fig. \ref{fig:grouping-effect},
it can be seen that for Shakeout, the units can be approximately separated
into two groups according to the maximum magnitudes of the connected
weights and the group around zero can be discarded, whereas
for Dropout, the units are concentrated. This implies that compared
to Dropout which may encourage a ``distributed code'' for the features
captured by the units of a layer, Shakeout tends to discard the useless
features (or units) and award the important ones. This experiment
result verifies the regularization properties of Shakeout and Dropout further.

As known to us, $L_{0}$ and $L_{1}$ regularization terms are related to performing
feature selection \cite{guyon2003introduction,7346492}. For a deep architecture, it is expected to obtain
a set of weights using Shakeout suitable for reflecting
the importance of connections between units.
We perform the following experiment to verify this effect. After a model is
trained, for the layer on which Dropout or Shakeout is applied, we
sort the magnitudes of the weights increasingly. Then we prune the
first $m\%$ of the sorted weights and evaluate the performance of the pruned model again. 
The pruning ratio
$m$ goes from 0 to 1. We calculate the relative accuracy loss (we
write $R.A.L$ for simplification) at each pruning ratio $m^{'}$
as
\[
R.A.L(m^{'})=\frac{Accu.(m=0)-Accu.(m^{'})}{Accu.(m=0)}
\]

Fig. \ref{fig:relative-accuracy-drop} shows the $R.A.L$
curves for Dropout and Shakeout based on 
the AlexNet model on ImageNet-2012 dataset. The models
trained by Dropout and Shakeout are under the optimal
hyper-parameter settings. Apparently, the relative accuracy loss for
Dropout is more severe than that for Shakeout.
For example, the largest margin of the relative accuracy losses between Dropout
and Shakeout is $22.50\%$, which occurs
at the weight pruning ratio $m=96\%$. 
This result proves that considering the trained weights in reflecting the importance of connections, Shakeout
is much better than Dropout, which benefits from the
implicitly induced $L_{0}$ and $L_{1}$ regularization effect. 
\setcounter{figure}{10}
\begin{figure*}
\centering

\subfloat[standard BP] {\includegraphics[bb=0bp 150bp 595bp 652bp,scale=0.28]{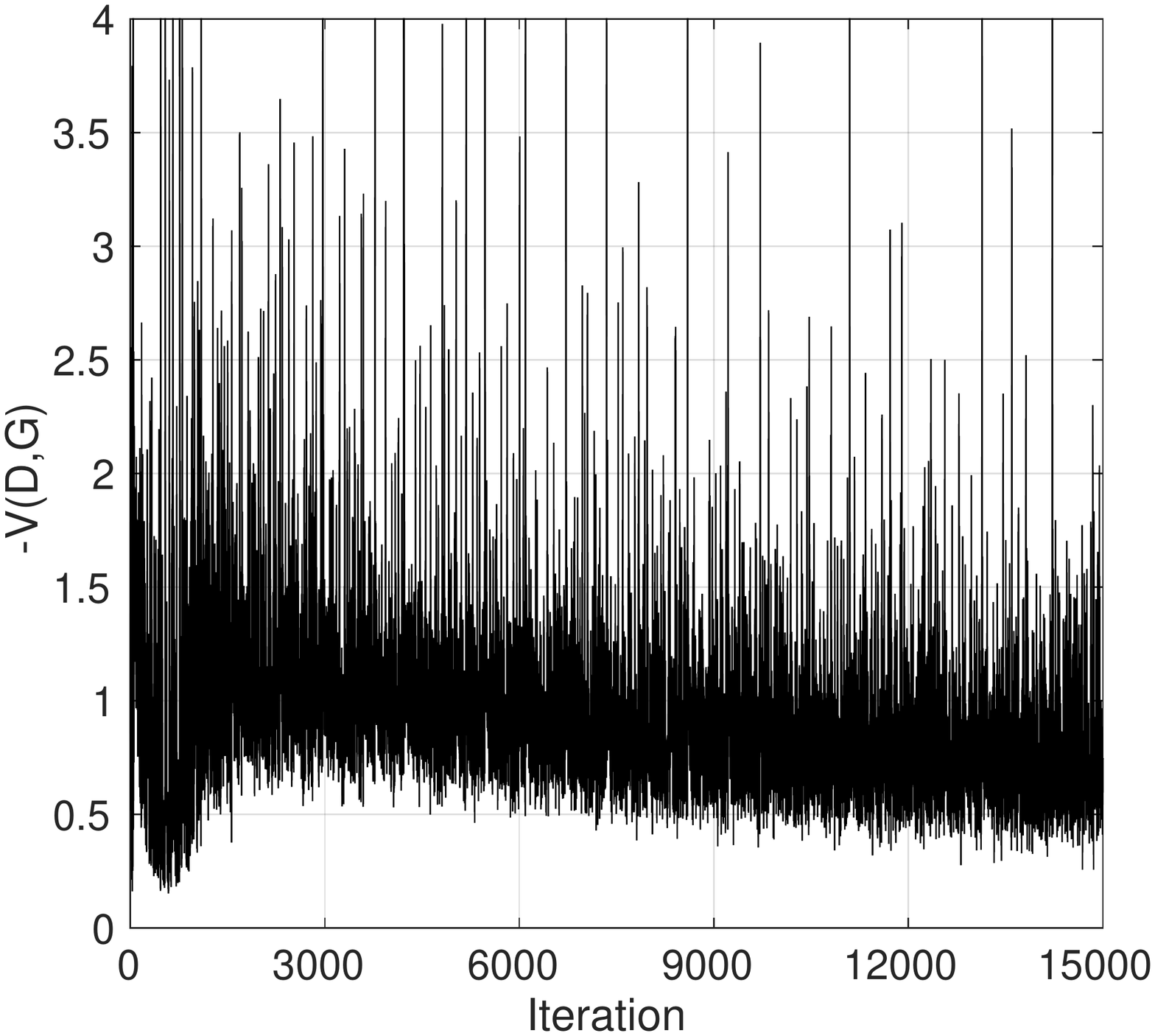}
\label{fig:DCGAN-objective-value-a}
}
\subfloat[Dropout] {\includegraphics[bb=0bp 150bp 595bp 652bp,scale=0.28]{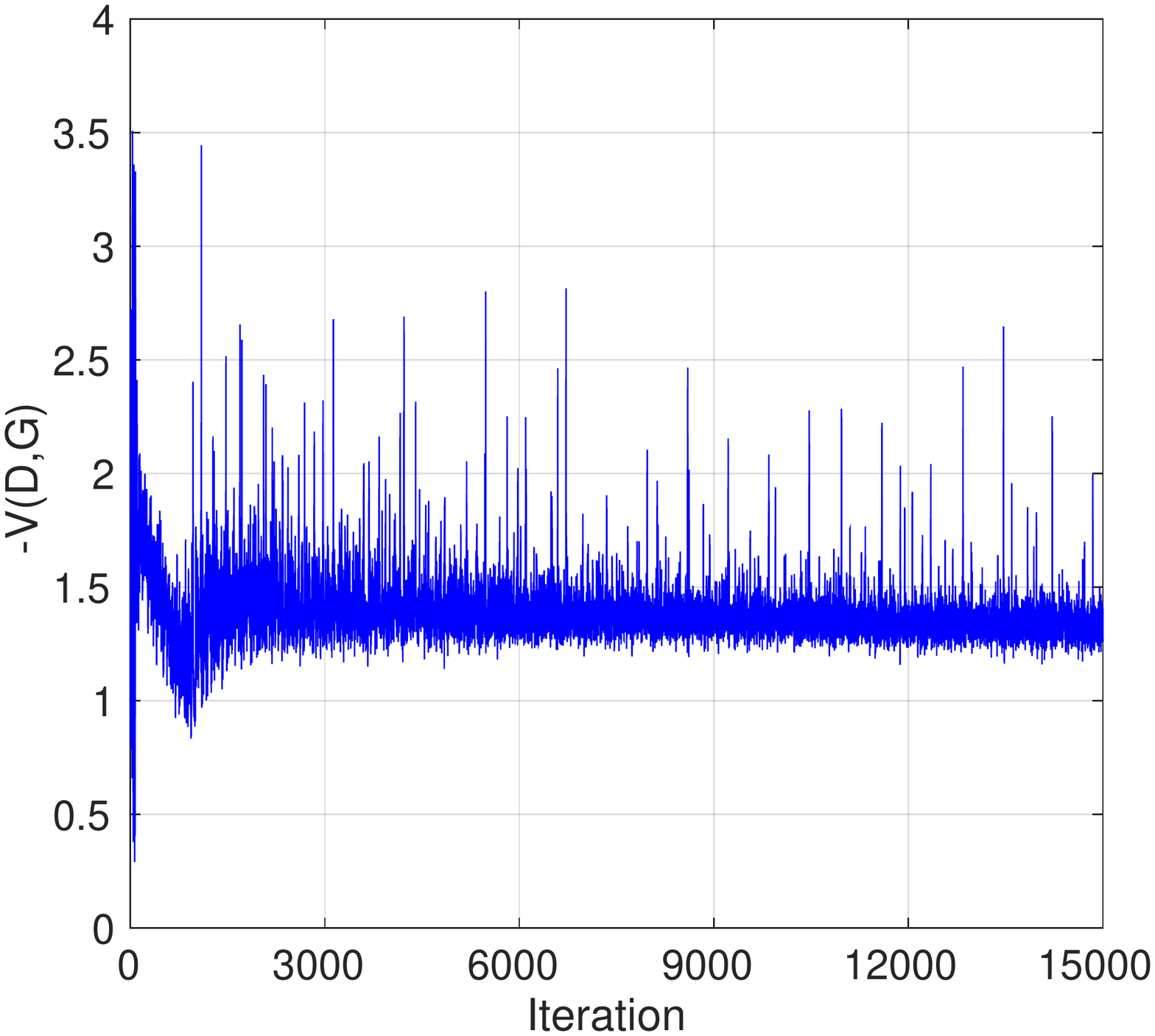}
\label{fig:DCGAN-objective-value-b}
}
\subfloat[Shakeout] {\includegraphics[bb=0bp 150bp 595bp 652bp,scale=0.28]{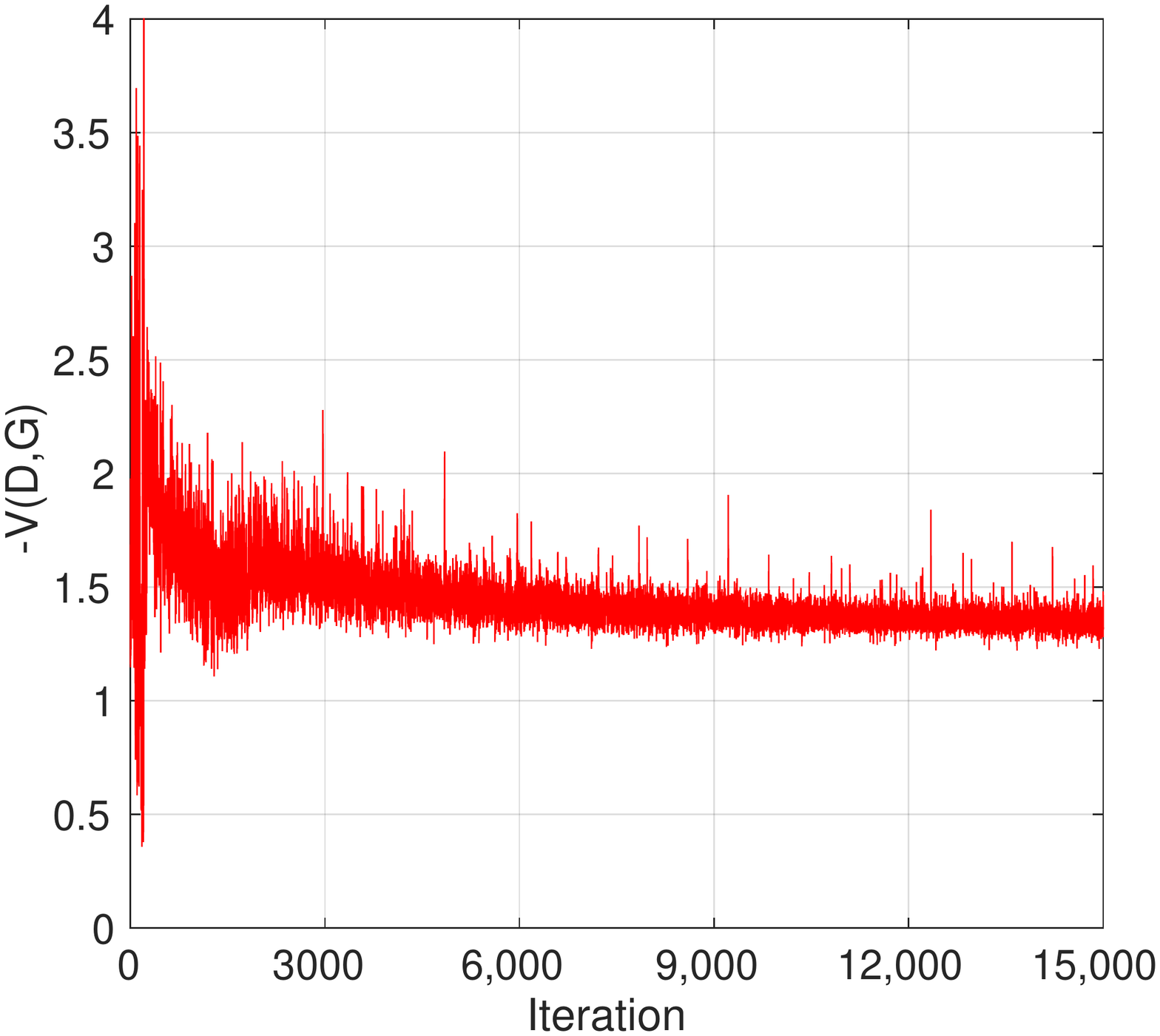}
\label{fig:DCGAN-objective-value-c}
}

\caption{The value of $-V(D,G)$ as a function
of iteration for the training process of DCGAN. DCGANs are trained
using standard BP, Dropout and Shakeout for comparison. Dropout or
Shakeout is applied on the discriminator of GAN.}
\label{fig:DCGAN-objective-value}
\end{figure*}
This kind of property is useful for the popular
compression task in deep learning area which aims to cut the connections
or throw units of a deep neural network to a maximum extent without
obvious loss of accuracy. 
The above experiments illustrate that Shakeout can play a considerable
role in selecting important connections, which is meaningful
for promoting the performance of a compression task. This is a potential
subject for the future research.

\setcounter{figure}{9}
\begin{figure}
\centering
\includegraphics[bb=30bp 165bp 575bp 685bp,scale=0.30]{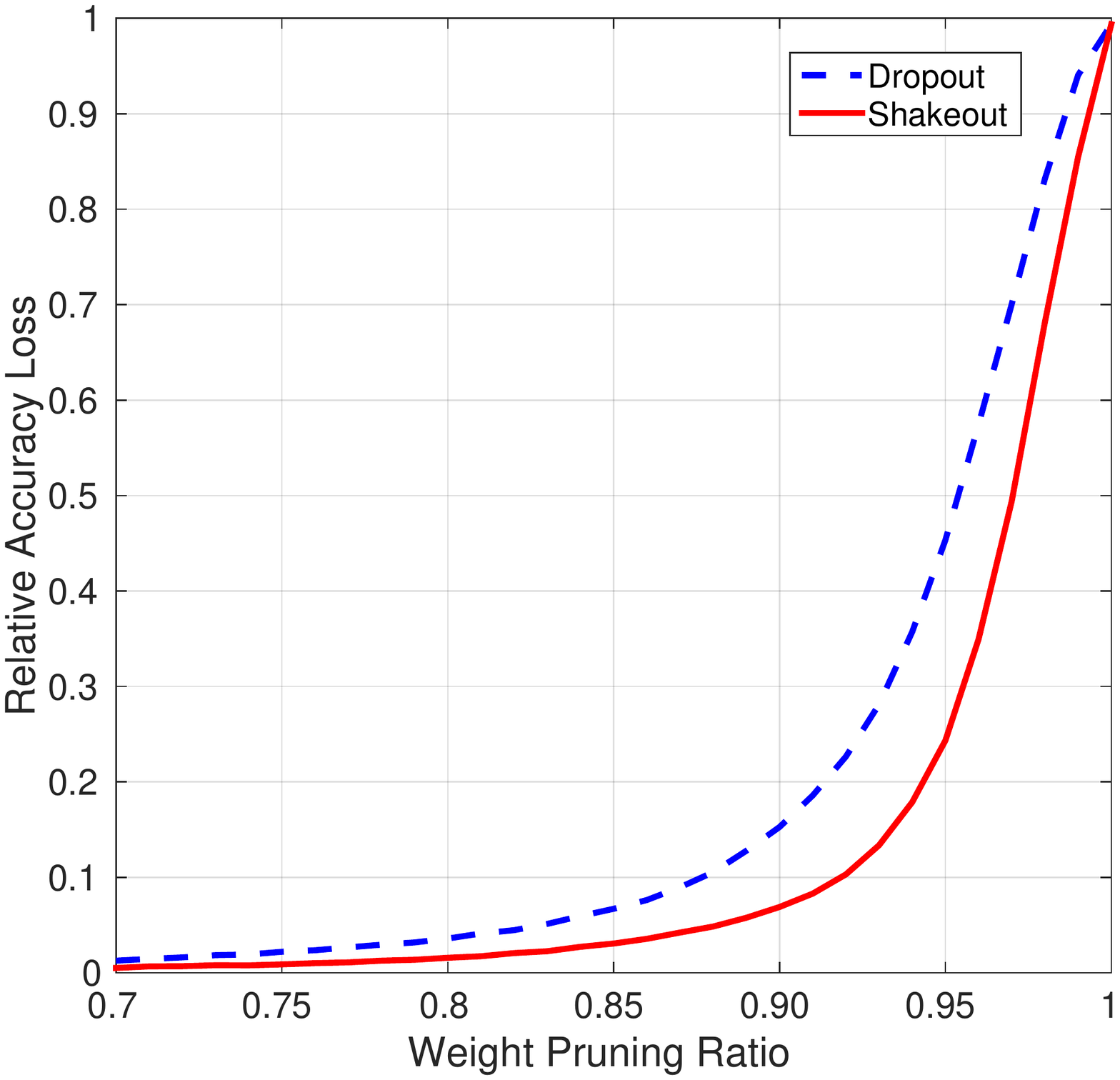}

\caption{Relative accuracy
loss as a function of the weight pruning ratio for Dropout and
Shakeout based on AlexNet architecture on ImageNet-2012. The relative accuracy loss for Dropout 
is much severe than that for Shakeout.
The largest margin of the relative accuracy losses between
Dropout and Shakeout is $22.50\%$, which occurs
at the weight pruning ratio $m=96\%$.}
\label{fig:relative-accuracy-drop}
\end{figure}
\setcounter{figure}{11}

\subsection{Stabilization Effect on the Training Process}
In both research and production, it is always desirable to have a
level of certainty about how a model\textquoteright s fitness to the
data improves over optimization iterations, namely, to have a \textit{stable}
training process. In this section, we show that Shakeout helps reduce
fluctuation in the improvement of model fitness during
training.

\begin{figure}
\centering
\includegraphics[bb=30bp 170bp 575bp 690bp,scale=0.32]{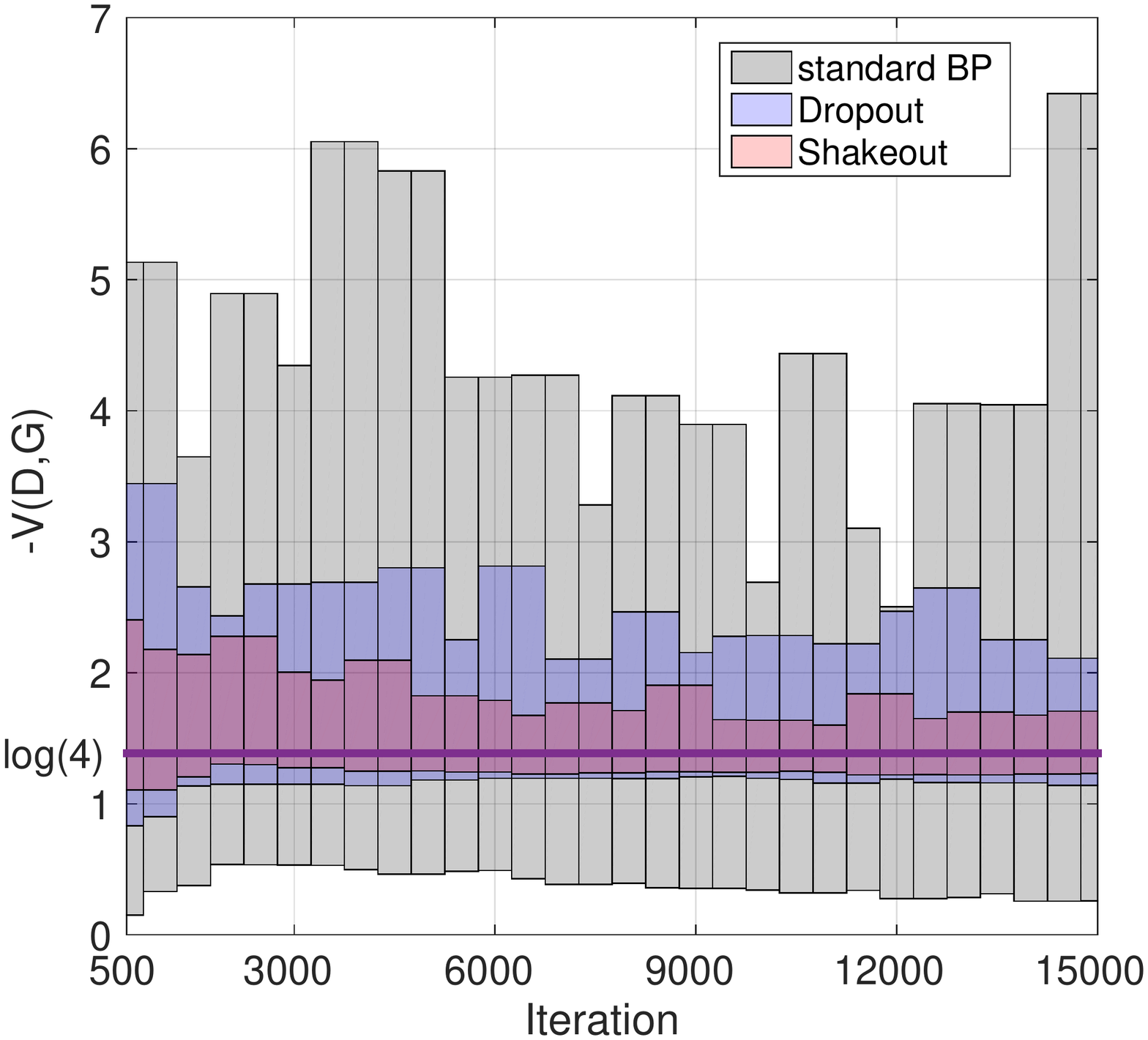}

\caption{The minimum and maximum values of $-V(D,G)$  within fixed length intervals moving from the start to the end of the training by standard BP, Dropout and Shakeout. The optimal value log(4) is obtained when the imaginary data distribution $P(\hat{\boldsymbol{x}})$ matches with the real data distribution $P(\boldsymbol{x})$.}
\label{fig:gan-minmax}
\end{figure}

The first experiment is on the family of Generative Adversarial Networks
(GANs) \cite{goodfellow2014generative}, which is known to be instable in the training stage \cite{radford2015unsupervised, arjovsky2017towards, arjovsky2017wasserstein}.
The purpose of the following tests is to demonstrate
the Shakeout\textquoteright s capability of stabilizing the training
process of neural networks in a general sense. 
GAN plays a min-max game between the generator $G$
and the discriminator $D$ over the expected log-likelihood of real
data $\boldsymbol{x}$ and imaginary data $\hat{\boldsymbol{x}}=G(\boldsymbol{z})$ where $\boldsymbol{z}$ represents the random input
\begin{equation}
\min_{G}\max_{D}V(D,G)=\mathbb{E}[\log[D(\boldsymbol{x})]
+\log[1-D(G(\boldsymbol{z}))]]
\end{equation}
The architecture that we adopt is DCGAN \cite{radford2015unsupervised}. 
The numbers of feature maps of the deconvolutional layers in the generator
are 1024, 64 and 1 respectively, with the corresponding spatial sizes
7$\times$7, 14$\times$14 and 28$\times$28. 
We train DCGANs
on MNIST dataset using standard BP, Dropout and Shakeout.
We follow the same experiment
protocol described in \cite{radford2015unsupervised} except for adopting
Dropout or Shakeout on all layers of the discriminator.
The values of $-V(D,G)$ during training are
illustrated in Fig. \ref{fig:DCGAN-objective-value}. It can be seen
that $-V(D,G)$ during training by standard BP oscillates
greatly, while for Dropout and Shakeout, the training processes are
much steadier. Compared with Dropout, the training process by Shakeout
has fewer spikes and is smoother. Fig. \ref{fig:gan-minmax} demonstrates the minimum and maximum values of $-V(D,G)$ within fixed length intervals moving from the start to the end of the training by standard BP, Dropout and Shakeout. It can be seen that the gaps between the minimum and maximum values of 
$-V(D,G)$ trained by Dropout and Shakeout are much smaller than that trained by standard BP, while that by Shakeout is the smallest, which implies the stability of the training process by Shakeout is the best.

\begin{figure}[!t]
\centering
\includegraphics[bb=0bp 160bp 630bp 700bp,scale=0.40]{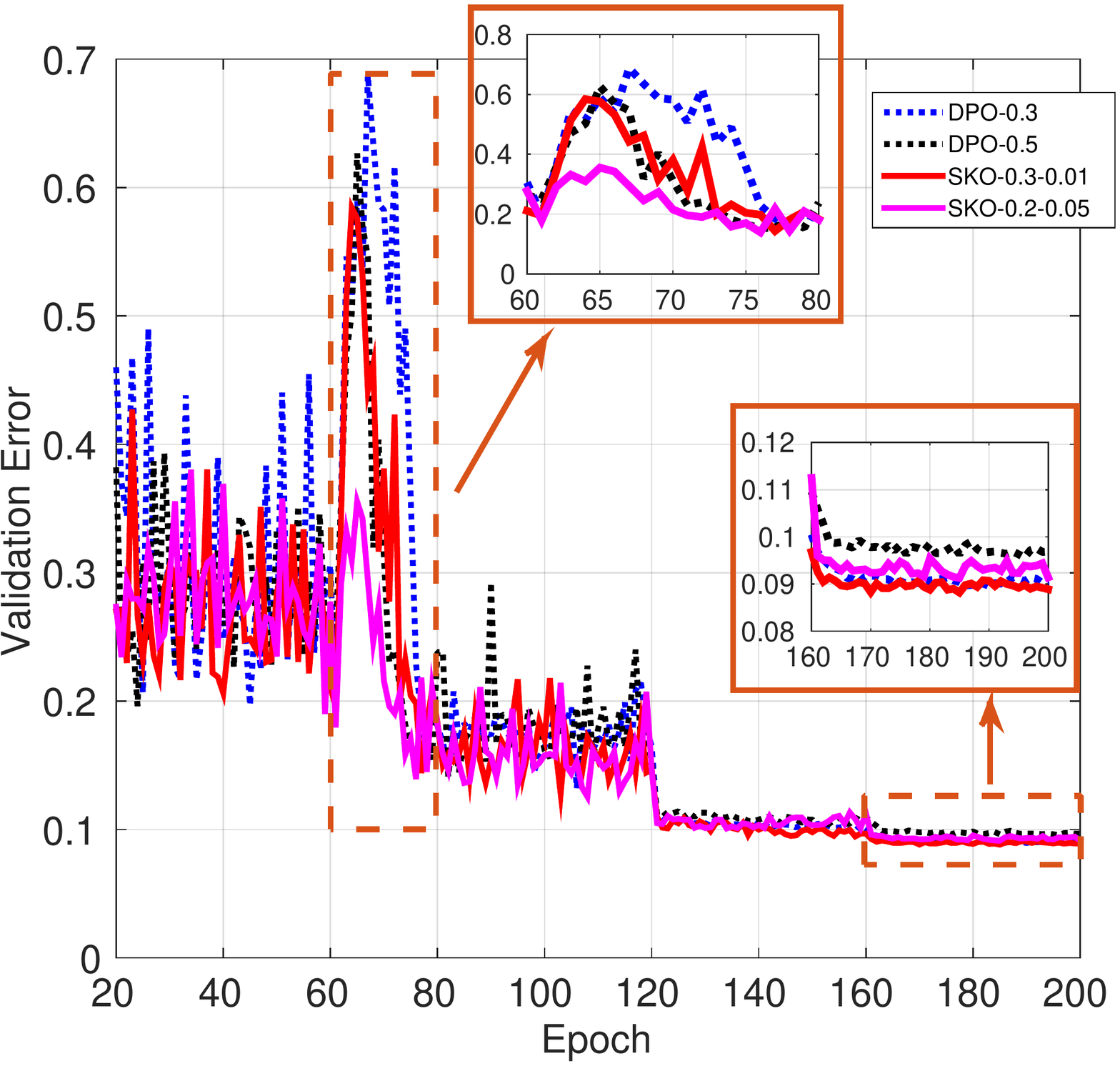}
\protect\caption{Validation error as a
function of training epoch for Dropout and Shakeout on CIFAR-10 with
training set size at 40000. The architecture adopted is WRN-16-4.
\textquotedblleft DPO"  
and \textquotedblleft SKO" 
represent \textquotedblleft Dropout" and 
\textquotedblleft Shakeout" respectively.
The following two numbers denote the hyper-parameters $\tau$ and $c$ respectively.
The learning rate decays at epoch 60, 120, and 160. 
After the first decay of learning rate, the
validation error increases greatly before the steady decrease (see the enlarged snapshot for training epochs from 60 to 80). It can
be seen that the extent of error increase is less severe for Shakeout than Dropout. Moreover, Shakeout recovers much faster than Dropout does. 
At the final stage, both of the validation errors steadily decrease (see the enlarged snapshot for training epochs from 160 to 200). 
Shakeout obtains comparable or even superior generalization performance to Dropout.}
\label{fig:CIFAR10-W16-4-training-curve}
\end{figure}

The second experiment is based on Wide
Residual Network architecture to perform the classification task. 
In the classification task, generalization
performance is the main focus and thus, we compare the validation errors
during the training processes by Dropout and Shakeout.
Fig. \ref{fig:CIFAR10-W16-4-training-curve}
demonstrates the validation error as a function of the training
epoch for Dropout and Shakeout on CIFAR-10 with 40000
training examples. The architecture adopted is WRN-16-4.
The experiment settings are the same as those described in Section \ref{sec:cifar-10-sec}.
Considering the generalization performance, the learning rate schedule adopted is 
the one optimized through validation to make the models obtain the best generalization performances.
Under this schedule, we find that the validation error temporarily
increases when lowering the learning rate at the early stage of training,
which has been repeatedly observed by \cite{DBLP:conf/bmvc/ZagoruykoK16}.  
Nevertheless, it can be seen from Fig. \ref{fig:CIFAR10-W16-4-training-curve} that 
the extent of error increase is less severe for Shakeout than Dropout.
Moreover, Shakeout recovers much faster than Dropout does. At the
final stage, both of the validation errors steadily decrease. Shakeout obtains comparable or even 
superior generalization performance to Dropout.
In a word, Shakeout significantly stabilizes the entire training process 
with superior generalization performance.

%

\subsection{Practical Recommendations}
\noindent \textbf{\textit{Selection of Hyper-parameters}} The most practical and popular way to perform hyper-parameter selection
is to partition the training data into a training set and a validation set to evaluate
the classification performance of different hyper-parameters on it.
Due to the expensive cost of time for training a deep neural network,
cross-validation is barely adopted. There exist many hyper-parameter
selection methods in the domain of deep learning, such as the grid
search, random search \cite{bergstra2012random}, Bayesian optimization
methods \cite{snoek2012practical}, gradient-based hyper-parameter
Optimization \cite{maclaurin2015gradient}, etc. For applying Shakeout
on a deep neural network, we need to decide two hyper-parameters $\tau$
and $c$. From the regularization perspective, we need to decide the
most suitable strength of regularization effect to obtain an optimal
trade-off between model bias and variance. We have pointed out that
in a unified framework, Dropout is a special case of Shakeout when
Shakeout hyper-parameter $c$ is set to zero. Empirically we find
that the optimal $\tau$ for Shakeout is not higher than that for Dropout. After determining the optimal $\tau$, keeping the order of magnitude of hyper parameter $c$ the same as $\sqrt{\frac{1}{N}}$
($N$ represents the number
of training samples) is an effective choice.
If you want to obtain a model with much sparser weights but
meanwhile with superior or comparable generalization performance to Dropout, a relatively
lower $\tau$ and larger $c$ for Shakeout always works.

\noindent \textbf{\textit{Shakeout combined
with Batch Normalization}} Batch Normalization
\cite{DBLP:conf/icml/IoffeS15} is the widely-adopted technique to promote
the optimization of the training process for a deep neural network.
In practice, combining Shakeout with Batch Normalization to train
a deep architecture is a good choice. For example, we observe that
the training of WRN-16-4 model on CIFAR-10 is slow to converge without
using Batch Normalization in the training. Moreover, the generalization
performance on the test set for Shakeout combined with Batch Normalization
always outperforms that for standard BP with Batch Normalization consistently
for quite a large margin, as illustrated in Tab. \ref{tab:w16-4-cifar10}.
These results imply the important role of Shakeout in reducing over-fitting
of a deep neural network.

\section{Conclusion}
We have proposed Shakeout, which is a new regularized training approach
for deep neural networks. The regularizer induced by Shakeout 
is proved to adaptively combine $L_{0}$,
$L_{1}$ and $L_{2}$ regularization terms. Empirically we find that

1) Compared to Dropout, Shakeout can afford much larger models. Or to say, when the data is scarce, Shakeout outperforms Dropout with a large margin.

2) Shakeout can obtain much sparser weights than Dropout with superior or comparable generalization performance of the model. While for Dropout, if one wants to obtain the same level of sparsity as that obtained by Shakeout, the model may bear a significant loss of accuracy.

3) Some deep architectures in nature may result in the instability of the training process, such as GANs, however, Shakeout can reduce this instability effectively.

In future, we want to put emphasis on the inductive
bias of Shakeout and attempt to apply Shakeout
to the compression task.


%

\ifCLASSOPTIONcompsoc
  \section*{Acknowledgments}
\else
  \section*{Acknowledgment}
\fi

This research is supported by Australian Research Council Projects (No. FT-130101457, DP-140102164 and LP-150100671).

\ifCLASSOPTIONcaptionsoff
  \newpage
\fi



\bibliographystyle{IEEEtran}
\bibliography{Kang}
\end{document}